\documentclass[12pt]{article}
\usepackage{float}%
\usepackage[affil-it]{authblk}
\usepackage{amssymb}
  \expandafter\let\csname equation*\endcsname\relax
  \expandafter\let\csname endequation*\endcsname\relax
\usepackage{amsmath}
\usepackage{fullpage}
\usepackage{graphicx}
\usepackage{epstopdf}
\newcommand{\im}[1]{\mathbf{#1}}
\newcommand{\hsp}{\hspace{.1in}}
\newcommand{\argmin}[1]{\underset{#1}{\operatorname{arg}\,\operatorname{min}}\;}

 \usepackage{chngcntr}
\counterwithout{figure}{section}

\begin{document}

\title{Stabilizing dual-energy X-ray computed tomography reconstructions using patch-based regularization}


\author{Brian H. Tracey and Eric L. Miller 
\thanks{Tufts University, Department of Electrical and Computer Engineering, 161 College Ave., Medford, MA 02155 USA.  Electronic address: \texttt{btracey@eecs.tufts.edu}}}

\date{}
\maketitle

\begin{abstract}
Recent years have seen growing interest in exploiting dual- and multi-energy measurements in computed tomography (CT) in order to characterize material properties as well as object shape.  Materials characterization is performed by decomposing the scene into constitutive basis functions, such as Compton scatter and photoelectric absorption functions. While well motivated physically, the joint recovery of the spatial distribution of photoelectric and Compton properties is severely complicated by the fact that the data are several orders of magnitude more sensitive to Compton scatter coefficients than to photoelectric absorption, so small errors in Compton estimates can create large artifacts in the photoelectric estimate.   To address these issues, we propose a model-based iterative approach which uses patch-based regularization terms to stabilize inversion of photoelectric coefficients, and 
solve the resulting problem though use of computationally attractive Alternating Direction Method of Multipliers (ADMM)  solution techniques.  Using simulations and experimental 
data acquired on a commercial scanner, we demonstrate that the proposed
processing
can lead to more stable material property estimates which should aid materials characterization in future dual- and multi-energy CT systems.
\end{abstract}

\section{Introduction}
\label{sec:intro}  

Typical Computed Tomography (CT) systems provide an image of the spatially varying X-ray absorption within the object being imaged.  While CT is a key imaging modality in  medical, security, and non-destructive testing applications, this approach provides minimal information on material properties.  In recent years there has been substantial interest in dual- or multi-energy CT systems, in which measurements are made either with differing X-ray energy spectra~\cite{ying,alvarez}, or using energy-resolving detectors~\cite{shikhaliev2008energy}.  By exploiting the energy-dependence of X-ray attenuation, these systems can provide additional, valuable information regarding object material properties~\cite{johnson2007material,engler1990review}.

For dual-energy CT processing, an important first question is the choice of basis functions used to represent the energy dependence of materials comprising the scene.  For medical imaging, the range of tissue or contrast agent properties are relatively constrained and well known.  Thus, several authors have used the attenuation profiles of constituent materials as basis functions~\cite{kalender1986,williamson2006,Lehmann1981}. In this paper we are mainly concerned with  airport luggage screening, where the materials being scanned may vary greatly.  In this application it is more natural to employ Compton and photoelectric coefficient basis functions, as this basis set captures the dominant X-ray scattering mechanisms in the energy range used for CT and therefore describes a wide variety of materials~\cite{ying}. The Compton/photoelectric decomposition was used in early dual energy work ~\cite{alvarez1976energy}, which performed a polynomial fit to decompose the two collected sinograms into separate Compton and photoelectric sinograms.  Recovery of the Compton and photoelectric coefficient images was then accomplished via filtered back projections (FBP) performed separately for each of these sinograms.  This general approach was later extended by Ying, Naidu and Crawford (YNC)~\cite{ying}, who also perform a pre-reconstruction decomposition of the data into Compton and photoelectric sinograms, followed by FBP reconstructions.  Rather than using a polynomial decomposition, YNC obtained improved decompositions by solving a constrained optimization problem for each point in the sinogram, with non-negativity constraints applied to all coefficients.  A key challenge in any Compton/photoelectric-based approach is that reconstruction of the photoelectric image is difficult as the data is inherently less sensitive to this coefficient (see~\cite{Semerci2012} for a detailed sensitivity analysis).  Thus,  YNC applies additional preprocessing steps to suppress noise in the estimated photoelectric sinogram.  The YNC method is regarded as the state-of-the-art in sinogram-decomposition based methods for dual energy luggage imaging~\cite{Semerci2012}.

Several authors have demonstrated that dual-energy imaging performance can be improved by applying iterative reconstruction methods, rather than FBP-based reconstructions~\cite{sukovic,de2001iterative,fessler&sukovic}.  Iterative methods (while computationally  more expensive) offer greater flexibility in modeling the system accurately and in accounting for noise statistics, and allow prior information to be incorporated in the solution process, generally in the form of regularization terms~\cite{kaipio2005statistical}.  These regularization terms may include, for example, L1-type penalties that encourage sparse solutions or Total Variation (TV) penalties that encourage piecewise constant solutions.  The TV penalty has been shown to reduce artifacts, especially for sparsely sampled CT data, in both monoenergetic CT reconstruction~\cite{tang2009performanceTV,LouTVnonlocal2010,SidkyPanTV2008,Huang2012_TVforsparseCT} and also in dual-energy reconstruction~\cite{minConeBeamTV,Zbijewski_dualEnergyTV,li2012_dualEnergyTV}, with the caution that TV can lead to overly smoothed or `patchy' images for low-SNR data~\cite{tang2009performanceTV}.  In addition, TV penalties must clearly be used with care during reconstruction of textured objects where piecewise constant solutions are not appropriate.  Thus there is a need for regularization approaches that can better adapt to the spatial structure of the scene being reconstructed. 

Such an alternative is potentially provided by patch-based regularization, which has attracted attention since the work of  Buades \emph{et al.}~\cite{BuadesNLMreg2006}.   The concept behind patch-based regularization is that in some applications, a stable  image is available which can be used as a reference to guide the inversion of a less stably estimated image with similar geometry.   A set of patch similarities is calculated from the reference image, which captures the location of objects and edges within the image.  These patch similarities are then used during recovery of the image of interest by applying an edge-preserving denoising algorithm such as non-local means (NLM)~\cite{Buades2005} to reduce the effects of noise on image inversion and thus stabilize the image recovery.  In  most applications, however, a reference image is not available.  Thus in~\cite{BuadesNLMreg2006}, the reference image was taken to be the estimate from the previous iteration of an iterative image restoration process.  This concept has been applied to mono-energetic iterative CT reconstruction~\cite{HuangNLMCT2011195} and PET imaging~\cite{wang2012}. 
 A modified version of patch-based regularization has been applied to \emph{multi-modal} reconstruction problems in CT-SPECT imaging~\cite{Chun2012}, with the CT image used as a reference image whose geometry can be used to stabilize the SPECT.  Because the CT and SPECT images may not both be sensitive to the same features in the image, ~\cite{Chun2012} includes a parameter which indicates how tightly the two images should be coupled.  

Our dual-energy iterative reconstruction approach is motivated by the observation that the stably estimated Compton image provides a natural reference for guiding reconstruction of the photoelectric image.  Because materials in luggage have both a Compton and a photoelectric signature, edges or objects present in one coefficient image should also be present in the other.  Thus unlike with the CT-SPECT problem ~\cite{Chun2012}, where there was a need to consider objects that might be visible in one modality but essentially invisible in the other, the dual-energy physics allows us to more tightly couple the two solutions.  Our group explored a similar regularization concept in \cite{Semerci2012}, in which we developed a regularization penalty that correlated edge maps of the Compton and photoelectric images.  Moving from edge map correlation to a patch-based regularization has two key advantages.   First, we are moving from a non-convex regularization term to a convex one (as discussed in detail below), which lets us exploit solution methods such as the Alternating Direction Methods of Multipliers (ADMM) approach.  Second, because the patch-based method imposes an edge-preserving smoothing, rather than directly encouraging the development of edges in the photoelectric image, it is much less likely to create artifacts in the photoelectric image whose edges correlate with the edges of Compton artifacts.  We demonstrate this last point in simulation results below.

A second observation is that, because photoelectric reconstruction is much more difficult than Compton reconstruction, it may be appropriate to apply different regularization strategies for Compton and photoelectric.  Thus, we demonstrate below a framework which applies TV regularization to the Compton image to reduce artifacts, while using patch-based regularization to stabilize the photoelectric image based on geometric structure learned from the recovered Compton image.
  
Our contributions in the present paper are:  1) we describe in detail the regularization strategy described in the previous paragraph, which combines TV and patch-based regularization; 2) we  applying the recently developed Alternating Direction Method of Multipliers (ADMM) techniques to the regularized dual-energy reconstruction problem, and 3) we demonstrate the above techniques on both simulated and experimental data, comparing against previous methods~\cite{ying,Semerci2012}. To obtain quantitative results from experimental data (where ground truth is not known), we tabulate the estimated material properties for a set of CT slices with varying clutter but with a set of objects (water, rubber sheets, etc.) which are known to be homogeneous and to have identical material properties.  Our proposed method leads to material parameter estimates with improved \emph{repeatability} across slices as well as improved \emph{homogeneity} within objects.  

The structure of this paper is as follows.  In Section 2, we review the physical model used for polyenergetic, dual-energy CT image formation, and describe the inverse problem being solved.  We then describe the unique features of the solution method, namely a) use of non-local means methods to regularize the photoelectric image and b) an ADMM-based solution technique.   Results of applying these methods to data are shown in Section 3, and we conclude in Section 4.


\section{Methods}

Typical X-ray sources used in CT applications generate an energy spectra roughly between $20$KeV and $140$KeV \cite{beutel2000handbook}.  In this energy range X-ray attenuation physics are dominated by Compton scatter and photoelectric absorption. We model these phenomena as a product of energy- and material-dependent terms \cite{alvarez} as follows 
\begin{equation}
\label{attenuation}
\mu(x,y,E) = c(x,y)f_{KN}(E)+p(x,y) f_p(E)
\end{equation}
where $\mu(x,y,E)$ is the total attenuation and $c(x,x)$ and $p(x,y)$ are the material dependent Compton scatter and photoelectric absorption coefficients respectively.  The quantity $f_{KN}$ is the Klein-Nishina cross section for Compton scattering which is given as:

\begin{equation}
\label{klein-nishina}
f_{KN}(\alpha) = \frac{1+\alpha}{\alpha^2}\left[\frac{2(1+\alpha)}{1+2\alpha}-\frac{1}{\alpha}\ln(1+2\alpha)\right] \quad  +\frac{1}{2\alpha}\ln(1+2\alpha)-\frac{1+3\alpha}{(1+2\alpha)^2}
\end{equation}
where $\alpha=E/510.95$KeV. Lastly, $f_p$ approximates the energy dependency of the photoelectric absorption and is given as $f_p = E^{-3}$.  The units of $\mu(x,y,E)$ and $c(x,x)$ are $cm^{-1}$, while $p(x,y)$ has units of $KeV cm^{-1}$. 


A dual energy CT system samples these images by measuring attenuation along ray-paths connecting multiple sources and detectors.  This measurement process is captured in a \emph{system matrix} $\im{A}$, which is the mapping from $N$ image pixels to $M$ CT ray paths.  The system matrix can be computed 
using ray-trace~\cite{siddon85} or projection methods~\cite{deman2004} given knowledge of the system geometry.  In either case, we assume that a single-energy scan yields a set of $M$ measurements. The process is repeated for low and high energy source spectra to create a modeled data vector $\mathbf{m}^T = [ \mathbf{m}_L^T, \mathbf{m}_H^T ]$ which consists of $2M$ elements.   The $i$th measurement for the low energy scan is  written  as:
\begin{equation}
\label{projectiondL}
[\mathbf{m}_L(\boldsymbol{\theta})]_i  = -\ln\frac{[\mathbf{Y}_L(\boldsymbol{\theta})]_i} {Y_{0,L}}
\end{equation}
and similarly for the high-energy scan data, $[\mathbf{m}_H]_i$.  Other processing steps (offset correction, gain correction, etc.) are also commonly applied to the acquired sinograms.  The set of all such measurements for one scan is known as a \emph{sinogram}.

We model the low- and high-energy measurements $\mathbf{Y}_L$ and $\mathbf{Y}_H$ as vectors of additive Poisson-Gaussian random variables given as
\begin{equation}
[\mathbf{Y}_L(\boldsymbol{\theta})]_i = \mbox{Poisson}\left\{[\overline{\mathbf{Y}}_L(\boldsymbol{\theta})]_i\right\} + \mbox{Normal}(0,\sigma_{e,L})
\end{equation}
with $[\mathbf{Y}_H(\boldsymbol{\theta})]_i$ being similar.  The Poisson variables account for the X-ray counting statistics, while the Gaussian terms captures detector electronics noise.  
The corresponding means $[\overline{\mathbf{Y}}_L]_i$ and $[\overline{\mathbf{Y}}_H]_i$ are given as
\begin{eqnarray}
\label{YdL} 
[\overline{\mathbf{Y}}_L]_i = \int S_L(E)\mathrm{exp}\bigl(-f_{KN}(E)\mathbf{A}_{i*}\mathbf{c}(\boldsymbol{\theta)} 
-f_{p}(E)\mathbf{A}_{i*}\mathbf{p}\boldsymbol({\theta})\bigr)\mathrm{d}E
\end{eqnarray}
and
\begin{eqnarray}
\label{YdH}
[\overline{\mathbf{Y}}_H]_i= \int S_H(E)\mathrm{exp}\bigl(-f_{KN}(E)\mathbf{A}_{i*}\mathbf{c} 
-f_{p}(E)\mathbf{A}_{i*}\mathbf{p}\bigr)\mathrm{d}E .
\end{eqnarray}
Here, $\mathbf{A}_{i*}$ is the i$^{\mbox{th}}$ row of $\mathbf{A}$, and $S_L(E)$ and $S_H(E)$ correspond to normalized low and high energy X-ray spectra.  Note that the data will also include scatter contributions not captured above.  However, we neglect scatter here, as scatter corrections  can be applied to measured data in pre-processing~\cite{Siewerdsen2006}. 

Our goal is estimation of the  Compton and photoelectric images.  While it is possible to represent the images using more coarse-grained basis functions~\cite{Semerci2012}, here we operate in the pixel domain.  Lexicographically unwrapping  the two images $\im{c}$ and $\im{p}$ into two vectors of length $M$,  we collect the  unknowns as:
 \begin{equation}
 \boldsymbol{\theta} = [\im{c}^T \im{p}^T]^T.
 \label{eq:thetaDef}
 \end{equation}
To invert for the desired images, we stack the dual-energy CT observations into a data vector $\mathbf{y} = [ \mathbf{y}_L^T, \mathbf{y}_H^T ]^T$).  We then  seek a solution that minimizes the following: 
 \begin{eqnarray}
\label{eq:optimizationBG}
\argmin{\boldsymbol{\theta}} F(\boldsymbol{\theta}) & = &\frac{1}{2} 
\left( \im{y}-\im{m}(\boldsymbol{\theta}) \right)\boldsymbol{\Sigma}
\left( \im{y}-\im{m}(\boldsymbol{\theta}) \right)^T  \\ \nonumber
& + & 
R_{TV}(\im{c}) + 
R_{NLM}(\im{p} \mid \im{c}). 
 \end{eqnarray}
In addition, we may seek to enforce non-negativity constraints on Compton and photoelectric images, as negative coefficients are non-physical.  Here, we enforce non-negativity contraints on the Compton image ($\im{c} \succeq 0$), relying on patch-based regularization  for stabilizing the photoelectric coefficients.

The three terms in Eq.~\ref{eq:optimizationBG} are derived as follows.  The first term is a weighted least squares data fidelity term, which was derived  by Sauer and Bouman~\cite{boumanSauer1996} as a quadratic approximation to the Poisson log-likelihood function, which has been shown to be useful in dual energy CT~\cite{sukovic}.  Following~\cite{boumanSauer1996}, we take the covariance matrix to be a diagonal matrix whose t erms are the number of counts detected, i.e. $\boldsymbol{\Sigma} = diag(\boldsymbol{m})$.  Physically, this can be interpreted as preferentially weighting measurements with high counts, as these measurements correspond to ray paths where attenuation is low and therefore have good signal-to-noise ratio.  

The second and third terms in Eq.~\ref{eq:optimizationBG} are regularization terms chosen to to stabilize the inversion.  $R_{TV}$ imposes a Total Variation (TV) penalty on the Compton coefficient image, encouraging solutions that are piece-wise constant, and is defined by summing over pixels $k$: 
\begin{equation}
R_{TV}(\im{c}) = \lambda_{TV} \sum_{k=1}^N \mid \im{D} \im{c} \mid. 
\label{eq:Rtv}
\end{equation}
where $\im{D}$ is $2N \times N$ difference matrix which computes derivatives in the $x$ and $y$ directions (so $\im{D} = [\im{D}_x \;\im{D}_y]$, where $\im{D}_x$ and $\im{D}_y$ are difference matrices in the $x$ and $y$ directions respectively). As noted above, TV regularization has been widely applied to CT inversion problems.
The $R_{NLM}$ term conditions the photoelectric estimate on the  Compton image estimated during the previous iteration, and is described in detail in the next section. We note that the model also depends on knowledge of the source spectrum and system matrix, although these are assumed known and therefore are not shown explicitly.   

An important note is that the data fidelity term in Eq.~\ref{eq:optimizationBG} is convex (while this term is more complex than those used in monoenergetic CT~\cite{ramaniFessler2012}, it is composed by a series of  convexity-preserving operations).  Thus if convex regularization terms are used, we are able to apply convex solution methods~\cite{boyd2004convex} as described in Section~\ref{sec:admm}.

\subsection{Patch-based regularization of photoelectric image}

Our group's previous work used an edge-correlation regularization term to stabilize the photoelectric image.  This term, written using a pixel-wise representation of the image, is given as~\cite{Semerci2012} 
\begin{eqnarray}
R_{edge}(\im{c},\im{p}) = \lambda_{edge} \left[\frac{\| \im{D}\im{c} \|_2^2 \| \im{D}\im{p} \|_2^2}{ [(\im{D}\im{c})^T  (\im{D}\im{p})]^2}  -1 \right]^2
\end{eqnarray}
This penalty term has the desirable property that it
decreases as the correlation of the gradients increases in negative or positive direction and vanishes when they are perfectly correlated or anti-correlated. However, it is not convex in $(\im{c},\im{p})$.

 Instead, here we apply a non-local means regularization approach, which helps to reduce noise artifacts in the recovered image by building a denoising step into the inversion.  
Preliminary work on  patch-based regularization was presented in a conference~\cite{TraceyMillerSPIE2013}, but did not include TV regularization or the ADMM formulation outlined below.  Similar to~\cite{BuadesNLMreg2006}, we  define the regularization term $R_{NLM}$ as: 
\begin{equation}
R_{nlm} =  \lambda_{NLM}  \int \left( p(x,y) - NL^{Ref}(p(x,y))  \right)^2 dx \; dy
\end{equation}
where $NL^{Ref}(p(x,y))$ represents non-local means averaging of the \emph{photoelectric} image.  The superscript \emph{'Ref'} denotes that the denoising weights are calculated from a reference image $I_{ref}$.  This regularization encourages the photoelectric estimate to converge towards a smoothed version of itself, but with smoothing done in an edge-preserving manner so that important image details are retained.
The discrete representation of this regularizer is found by summing over pixels $k$
\begin{eqnarray}
\label{eq:nlmdiscrete}
R_{nlm} &  = &   \lambda_{NLM}  \sum_k \left( p(k) - NL^{Ref}( \im{p}(k)) \right)^2  \equiv \lambda_{NLM} \sum_k \delta_k^2
\end{eqnarray}
where $\delta_k$ has been defined as pixels in the `difference image', i.e. the difference between the smoothed and unsmoothed solutions.

The smoothed image $NL^{Ref}(\im{p})$ is found using the NLM method developed in the image denoising literature~\cite{WuJS2012,BuadesNLMreg2006}.   Classical denoising  suppresses noise by convolving the image using a data-independent kernel (for example the Gaussian kernel). In contrast, methods such as NLM use data-adaptive (and spatially varying) weights to achieve more accurate denoising results~\cite{Buades2005}.  In this approach, each pixel location $k$ is associated with a patch, typically a square region centered on the pixel.  The NLM kernel calculates the weights based on measures of  similarity between patches.  The most common similarity measure is mean-squared patch differences ~\cite{DeVille2009_SURE}, for which the NLM kernel becomes:

\begin{eqnarray}
K_{NLM}(k,l,\im{I}_{ref}) &=& \exp\left(-\frac{\sum_{\delta \in \Delta} (I_{ref}(k+\delta) - I_{ref}(l+\delta))^2}{2 \it{L_\Delta} \beta^2}  \right)  
\label{eq:3}
\end{eqnarray}
(note~\cite{Buades2005_rev} defined the denominator above as $h^2$).
In Eq.~\ref{eq:3}, $\beta$ is a bandwidth parameter, while $\Delta$ represents a local patch of pixels surrounding $k$, containing $\it{L_\Delta}$ pixels; a patch of the same shape also surrounds $l $, and $\delta$ indicates the offset from each patch center. Although a variety of patch shapes are possible~\cite{Buades2005_rev,Deledalle2011}, square patches centered on the points of interest are most common. 
If similar patches can be found throughout the image, then ideally the neighborhood $\textit{N}_{srch}$ is taken to be the entire image, so the averaging process is fully \emph{non-local}.  In practice, $\textit{N}_{srch}$ is usually limited to reduce computational load.   For example,~\cite{Buades2005} chose $7 \times 7$ patches for $\Delta$ and a $21\times 21$ pixel  search region.
The key NLM parameters are the patch size, specified as a half-width $W$ (so $\it{L}_\Delta = (2W+1)^2$), the size of $\textit{N}_{srch}$,  specified as a half-width $M$, and the bandwidth $\beta$.   While the results are most sensitive to the bandwidth parameter $\beta$, it is also important that the patch size be comparable to the smallest features of interest.

In the dual-energy problem, several choices for the reference image $\im{I}_{ref}$ are possible.  Filtered back-projection (FPB) reconstructions of single-energy scans provide computationally cheap reconstructions of image geometry, so one choice is to take $\im{I}_{ref} = \im{I}_{FBP}$, where the FBP reconstruction$ \im{I}_{FBP}$ can be computed from either the high- or low-energy scan.  With this choice, the smoothed photoelectric image in Eq.~\ref{eq:nlmdiscrete} is explicitly given as
\begin{equation}
NL^{FBP}(\im{p}(k)) =  {\sum_{l\in \textit{N}_{srch}(k) }K_{NLM}(k,l,\im{I}_{FBP}) p(k)\over \sum_{l\in \textit{N}_{srch}(k) }K_{NLM}(k,l,\im{I}_{FBP})}
\end{equation} 
which is clearly convex in $\im{p}$.  In our experimental results below, we instead use the \emph{Compton} image from the \emph{previous} iteration as a reference, so $\im{I}_{ref}^j = \im{c}_{prev}$ and the smoothing term from   Eq.~\ref{eq:nlmdiscrete} becomes
\begin{equation}
NL^{\im{c}}(\im{p}(k)) =  {\sum_{l\in \textit{N}_{srch}(k) }K_{NLM}(k,l,\im{c}_{prev}) p(k)\over \sum_{l\in \textit{N}_{srch}(k) }K_{NLM}(k,l,\im{c}_{prev})}
\end{equation} 
which is convex in $\im{p}$, and $\im{c}^{prev}$ is  regarded as fixed.  This approach (taking the previous iteration's estimate as the reference for non-local regularization) has been used in previous work~\cite{zhang2010_bregnl,chen2008_nlBaysianPrior,Adluru2010_nlMRI,Liang2011_nlReg}, although as pointed out in~\cite{wang2012} the approach is somewhat empirical and convergence proofs have not been established.  In our problem, we observed that the Compton image converges rapidly to a solution which closely mirrors the geometry of FBP, except for cases such as that examined in Fig.~\ref{fig:HCsubsampCompton} where the raypaths are sparsely sampled, making the choice of $\im{I}_{ref}$ less critical when sparse sampling is avoided. We use the integral image technique introduced in~\cite{Darbon2008} to reduce computation of NLM weights. 

%

\subsection{ADMM reformulation}
\label{sec:admm}

Here we present an ADMM approach to dual energy-reconstruction, which is similar to previous approaches for monoenergetic CT~\cite{ramaniFessler2012} but has been expanded  by introducing separate equations that solve for Compton and photoelectric coefficients, and by introducing nonnegativity constraints.
The problem in Eq.~\ref{eq:optimizationBG} can be reformulated as follows:  we define  $\im{z} = [\im{t}^T \im{u}^T  \im{v}^T \im{s}^T]^T$ as a vector of auxiliary constrained variables. These are related to the original parameter vector $\im{\theta}$ (from Eq.~\ref{eq:thetaDef}) as $\im{z} = \im{C} \im{\theta}$, where $\im{C} = [\im{I}_N \; \im{0}_N; \; \im{0}_N \; \im{I}_N; \; \im{D} \; \im{0}_D; \; \im{I}_N \; \im{0}_N]$, where $\im{I}_N$ is an $N \times N$ identity matrix and $\im{0}_N$ is an $N \times N$ matrix of zeros,
$\im{D}$ is the TV difference matrix from Eq.~\ref{eq:Rtv}, and $\im{0}_D$ is a vector of zeros of the dimension of $\im{D}$.  
Then, the overall problem can be written as:
\begin{eqnarray}
\label{eq:reform1}
\argmin{\im{z}} f(\im{z})& = &\frac{1}{2} \Vert \im{y} - F(\im{t},\im{u}) \Vert^2_{\im{W}} \\ \nonumber
& + & \lambda_{TV} \sum_{k=1}^N \mid \im{v} \mid + \Psi_{NLM}(\im{u} \mid \im{t})  + g(\im{s}) 
\end{eqnarray}
subject to
\begin{eqnarray}
\label{eq:constraints}
\im{z} = \im{C} \im{\theta}.  
\end{eqnarray}
where $g(\im{s})$ is an indicator function for the non-negative orthant.  Thus, $\im{t}$ and $\im{u}$ are auxiliary variables for the Compton and photoelectric images respectively, $\im{v}$ is related to the TV constraint, and $\im{s}$ captures non-negativity constraints on the Compton image.
Because $\im{C}$ is full column rank and $f(\im{z})$ is convex, Theorem 1 of~\cite{ramaniFessler2012} states that the solution to Eq.~\ref{eq:reform1}-\ref{eq:constraints} will converge to the solution of Eq.~\ref{eq:optimizationBG}.
The next step is forming the Augmented Lagrangian (AL)~\cite{boyd2011distributed}:
\begin{eqnarray}
\mathcal{L}(\im{\theta},\im{z},\im{\gamma},\mu) & \triangleq & f(\im{z}) + \im{\gamma}^T \im{\Lambda} (\im{z} - \im{C} \im{\theta}) + \frac{\mu}{2} \Vert \im{z} - \im{C} \im{\theta}\Vert_{\Lambda^2}
\end{eqnarray}
where the second term is the Lagrangian term, and the last term is an augmented term which is known to improve the convergence of the problem.  

One difference of the equation above from the standard AL discussed in \cite{boyd2011distributed} is the introduction of $\Lambda$.  This matrix is introduced by Ramani and Fessler~\cite{ramaniFessler2012} to 'balance' the sub-matrices in $\im{C}$, because the terms $\im{D}$ may be orders of magnitude different from unity.  In our case this  term would be $\im{\Lambda} = [\im{I}_{N} \hsp \im{0}_{N} \hsp \im{0}_{N}; \im{0}_{N} \hsp \im{I}_{N} \hsp \im{0}_{N}; \im{0}_{N} \hsp \im{0}_{N} \hsp \sqrt{\nu}\im{I}_{N}]$ where  $\nu$ is a scaling constant.  Ramani and Fessler state that $\nu$ and also $\gamma$ affect only the convergence rate, not the final answer, and give guidelines for choosing them.

Again following~\cite{ramaniFessler2012}, this scaling matrix can be absorbed into the quadratic penalty by defining $\boldsymbol{\eta}=\frac{1}{2} \im{\Lambda}^{-1} \im{\gamma}$. 
Then, the AL becomes (dropping constant terms that do not affect the solution)
\begin{eqnarray}
\mathcal{L}(\im{\theta},\im{z},\im{\gamma},\mu) = f(\im{z})  + \frac{\mu}{2} \Vert \im{z} - \im{C} \im{\theta}  -\boldsymbol{\eta} \Vert^2_{\Lambda^2}
\label{eq:finalL}
\end{eqnarray}
For convenience, the vector $\boldsymbol{\eta}$ can be written as $\boldsymbol{\eta} = [\boldsymbol{\eta}_{t}^T \hsp \boldsymbol{\eta}_{u}^T \hsp \boldsymbol{\eta}_{v}^T \hsp \boldsymbol{\eta}_{s}^T]^T$ to emphasize the terms associated with each auxiliary variable.

Finally, we use the ADMM setup to solve the problem using an alternating minimization scheme:
\begin{eqnarray}
\im{t}^{(j+1)} & = & \argmin{\im{t}}  \mathcal{L}(\im{\theta}^{(j)},\im{t}^{(j)},\im{u}^{(j)},\im{v}^{(j)},\im{s}^{(j)},\im{\gamma}^{(j)},\mu) \label{eq:tupdate} \\
\im{v}^{(j+1)} & = & \argmin{\im{v}}   \mathcal{L}(\im{\theta}^{(j)},\im{t}^{(j+1)},\im{u}^{(j)},\im{v}^{(j)},\im{s}^{(j)},\im{\gamma}^{(j)},\mu) \label{eq:vupdate} \\ 
\im{s}^{(j+1)} & = & \argmin{\im{s}}   \mathcal{L}(\im{\theta}^{(j)},\im{t}^{(j+1)},\im{u}^{(j)},\im{v}^{(j+1)},\im{s}^{(j)},\im{\gamma}^{(j)},\mu) \label{eq:supdate} \\ 
\im{u}^{(j+1)} & = & \argmin{\im{u}}  \mathcal{L}(\im{\theta}^{(j)},\im{t}^{(j+1)},\im{u}^{(j)},\im{v}^{(j+1)},\im{s}^{(j+1)},\im{\gamma}^{(j)},\mu) \label{eq:uupdate} \\ 
\im{\theta}^{(j+1)} & = & \argmin{\im{\theta}}   \mathcal{L}(\im{\theta}^{(j)},\im{t}^{(j+1)},\im{u}^{(j+1)},\im{v}^{(j+1)},\im{s}^{(j+1)},\im{\gamma}^{(j)},\mu)  \label{eq:xupdate}\\ 
\boldsymbol{\eta}^{(j+1)} & = & \boldsymbol{\eta}^{(j)} + (\im{z}^{(j+1)}-\im{C} \im{\theta}^{(j+1)})   \label{eq:etaupdate}\
\end{eqnarray}
Let us consider these problems one at a time.  Eq.~\ref{eq:tupdate} updates the \emph{Compton image coefficients}.  Solving this first is a natural choice, as the Compton image is the most stable.  Dropping constant terms, we seek to minimize:
\begin{equation}
\argmin{\im{t}^{(j+1)}}
\frac{1}{2} \Vert \im{y} - F(\im{t}^{(j+1)},\im{u}^{(j)}) \Vert^2_{\im{W}} + \frac{\mu}{2} \Vert \im{t}^{(j+1)} - \im{c}^{(j)}  -\boldsymbol{\eta}^{(j)}_{t} \Vert^2
\end{equation}
where the constraint terms (2nd term) are simplified because the weights in $\im{\Lambda}$ are 1 for the terms in $\im{t}$, and we can use the fact that $\im{c}^{(j)} = \im{\theta}^{(j)}(1:N_R) $). This problem is a weighted non-linear least-squares problem and can be solved using standard approaches~\cite{Semerci2012}, projecting the result into the non-negative orthant.  In doing so, we need to calculate the derivative terms $\frac{\partial \im{J}}{\partial t_i}$. This derivative is the derivative of the data fidelity terms (calculated in ~\cite{Semerci2012}) and the derivative of the augmented Lagrangian.  The AL derivative terms gives a contribution
\begin{equation}
\mu \sum_k  \left( t(k)^{(j+1)} - c(k)^{(j)}  - \eta(k)^{(j)}_{t} \right)
\end{equation}
which is added to every element of $\frac{\partial \im{J}}{\partial t_i}$.

Next,  Eq.~\ref{eq:vupdate} seeks to minimize the sum of the  \emph{Total Variation} term and the relevant terms in the $\Vert \Vert_{\Lambda^2}$ term in Eq.\ref{eq:finalL}. As shown in Boyd et al.~\cite{boyd2011distributed}, it is solved by soft thresholding, on a term-by-term basis:
\begin{equation}
\im{v}^{(j+1)} = S_{\lambda_{TV}/(\mu \nu)} (\im{D} \im{c}^{(j)} + \boldsymbol{\eta}_v)
\end{equation}
The argument of the softmax function $S_{\lambda_{TV}/(\mu \nu)} $ is found by setting the constraint term ($\im{z} - \im{C} \im{\theta}  -\boldsymbol{\eta}$) to zero and solving for $\im{v}$ (ignoring the other terms in $\im{z}$).  A similarly simple solution is found for Eq.~\ref{eq:supdate}, which imposes the non-negativity constraint by projecting onto the non-negative orthant~\cite{boyd2011distributed}:
\begin{equation}
\im{s}^{(j+1)} = max(0,\im{c}^{(j)} + \boldsymbol{\eta}_s)
\end{equation}

Next, Eq.~\ref{eq:uupdate} updates the  \emph{photoelectric image coefficients}.  It solves a  similar problem to that addressed when solving the Compton coefficients:
\begin{equation}
\argmin{\im{u}^{(j+1)}}
\frac{1}{2} \Vert \im{y} - F(\im{t}^{(j+1)},\im{u}^{(j+1)}) \Vert^2_{\im{W}} + \Psi_{NLM}(\im{u}^{(j+1)} \mid \im{t}^{(j+1)}) + \frac{\mu}{2} \Vert \im{u}^{(j+1)} - \im{p}^{(j)}  -\boldsymbol{\eta}^{(j)}_{u} \Vert^2
\end{equation}
The key difference is that this problem involves the NLM regularization term; here, the weights in NLM are calculated from the previous iteration's Compton image.  
This term's contribution to the Jacobian can be shown to be~\cite{TraceyMillerSPIE2013} 
\begin{eqnarray}
J_{nlm}(i)  = \frac{\partial R_{nlm}}{\partial \im{p}(i)} = 2  \left( \delta_i  - \sum_k \delta_k \frac{w^{(C)}_{ik}}{Z^{(C)}_k}) \right)
\end{eqnarray}
Here, note that the second term is equal to a NLM-smoothed version of the difference image $\delta_k$ (defined in  Eq.~\ref{eq:nlmdiscrete}).  Thus, it can easily be found by calling the NLM code twice, using the Compton image from the previous iteration in both cases to calculate weights.  Because we use the previous iteration's Compton image, $R_{NLM}$ does not depend on the current Compton estimate, so $\partial R_{nlm} /  \partial \im{c}=0$.  

Like Eq.~\ref{eq:tupdate}, this can be solved using Levenberg-Marquart.  Here, the augmented Lagrangian adds a term of 
\begin{equation}
-\mu \sum_k  \left( u(k)^{(j+1)} - \im{p}(k)^{(j)}  - \eta(k)^{(j)}_{t} \right)
\end{equation}
to every element of $\frac{\partial \im{J}}{\partial u_i}$, as well as adding to the error.

Eq.~\ref{eq:xupdate} applies the constraints to adjust $\im{\theta}$ into agreement with the auxiliary variables.  The problem to be solved is:
\begin{equation}
\argmin{\im{\theta}^{(j+1)}}
 \Vert \im{z}^{(j+1)} - \im{C} \im{\theta}^{(j+1)}   -\boldsymbol{\eta} \Vert^2_{\Lambda^2}
\end{equation}
which is a weighted least squares problem.  

Finally, Equation~\ref{eq:etaupdate} updates the Lagrange multipliers.  The solution continues iteratively  until a preset number of iterations is exceeded, or the primal and dual residuals fall below a  threshold (see~\cite{boyd2011distributed} for details).

During numerical testing, we found the TV term to require many iterations to converge.  However, a single iteration of the TV term is very cheap to calculate, as is a single iteration of the non-negativity term, whereas a single iteration of the Compton or photoelectric terms is very expensive.  We therefore modify the approach above to complete many TV and non-negativity iterations after each Compton update.  The terms involving $\im{v}$ and $\im{s}$ in Eq.~\ref{eq:xupdate} and \ref{eq:etaupdate} are updated while other parameters are held constant.



\textit{Computation}: Because our problem sizes are much larger than those examined in~\cite{oguzThesis,Semerci2012}, explicit storage of $\im{J}$ and $\im{J}^T \im{J}$ is not possible.  Similar to other authors~\cite{NargolThesis}, we reduce memory storage by computing $\im{J}^T \im{J} \im{\theta}$ in two steps (first, $\im{y}=\im{J} \im{\theta}$, then $\im{J}^T \im{y}$).  We also reduce memory requirements by reformulating the Jacobian terms as products of stored vectors with the system matrix, then computing $\im{J}$ and $\im{J}^T$ on-the-fly as needed.   
Because we are computing in a Linux cluster environment and are assigned compute nodes based on current availability, the run times for our solutions vary considerably depending on the node used.  However, typical run times for a single slice are between 6-9 hours.   The computational cost of regularization (both TV and NLM-based regularization) is an increase in run time of roughly 20\%.   
Most of the computational effort comes in solving the conjugate gradient (CG) problem that lies at the heart of the Levenberg-Marquart solver. Recently an approach to subsampling the ray paths used in the CG solver has been developed~\cite{LLNLCG}, giving roughly $\sim20 \times$ speed up.  While we were not aware of this approach until our code was fully developed, incorporating this approach into our framework should give noticeable computational savings.

\section{Results}

Here we present processing results using the method outlined above. Our main focus is on experimental data collected on the Imatron C300 CT scanner, though we also characterize algorithm performance on simulated data created to mimic the experimental geometry. We therefore first briefly discuss the Imatron system geometry and energy spectra used, then discuss parameter selection and implementation of the legacy YNC. We then present results on simulated and experimental data, demonstrating qualitative improvements in image quality and quantitative improvements in image metrics.

\subsection{Description of Imatron C300 system}

Experimental data were acquired from a set of dual-energy scans performed on the Imatron C300 CT scanner.  This commercial single-energy scanner (developed for the cardiology market) was re-purposed for dual-energy scans by performing sequential scans of each object, with X-ray source voltage adjusted between scans.  In all cases here, the registration between the sequential scans appears excellent.  The Imatron scanner performs helical scans, with the source scanning through 2588 source positions covering 210 degrees, with measurements made on 864 detector channels.  The Imatron system software can be used to output a wide variety of intermediate results, ranging from raw data to single-energy FPB reconstructions.   While the regularization methods we present are applicable to general geometries, we chose to work with sinograms where data was rebinned into a parallel beam geometry, with scatter corrections  applied to the data.   

\begin{figure}
\centering
\includegraphics[width=6 cm]{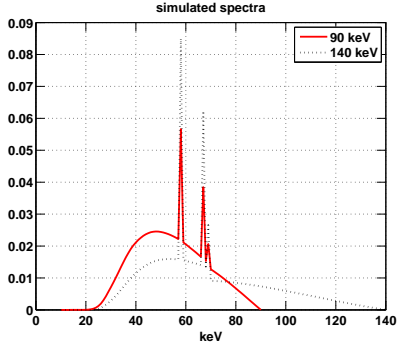}
\caption{Normalized X-ray energy spectra for high- and low-energy scans}
\label{fig:spectra}
\end{figure}

One disadvantage of using a standard scanner to acquire dual-energy data is that the X-ray source energies cannot be tuned over as wide a range as would be typical for a dedicated dual-energy system.  Figure~\ref{fig:spectra} shows the estimated high- and low-energy spectra for our data, modeled based on detector settings.  The significant overlap between the spectra makes the decomposition problem somewhat more ill-posed that it would be if scan energies were better separated.

\subsection{Parameter selection and legacy method comparison}
The regularization approach outlined above requires selection of several parameters including the weights for TV and NLM penalties ($\lambda_{TV}$ and $\lambda_{NLM}$ and denoising parameters for the NLM term (bandwidth $\beta$, patch size, and search neighborhood).  Methods for optimizing regularization parameter selection exist, including L-curve and Generalized Cross-Validation (GCV), as discussed in~\cite{Semerci2012}. NLM denoising parameter selection can be approached by minimizing Stein's Unbiased Risk Estimate (SURE)~\cite{DeVille2009_SURE}, potentially in combination with additional constraints~\cite{Tracey2013ConstrOptNLM}.  However, the above-discussed methods are not straightforward to apply.  Because our present goal is a proof of concept that our approach offers potential benefits for dual-energy reconstruction, we instead manually tuned parameters based on visual inspection of the results. NLM parameters were selected to give good denoising of typical Compton images from our dataset (i.e., results that were visually judged to reduce noise without smearing small object details).  The motivation for this approach is that parameters which create weights that lead to good Compton denoising  should successfully capture image structure, and are therefore also appropriate for photoelectric regularization.  We found $\beta=0.5e^{-4}$, 7 pixel $\times$ 7 pixel patches, and  19 pixel $\times$ 19 pixel search neighborhoods to be a good choice across the images examined.   In addition, because many objects in our experimental dataset have texture, we set the TV penalty weight to be low ($\lambda_{TV} = 0.01$).  We used these values across all images instead of re-tuning to optimize individual images.

To evaluate the quality of our results, we compare against an implementation of the YNC dual-energy reconstruction method proposed by Ying \emph{et al.}~\cite{ying}. 
The version of YNC \emph{et al.} implemented here includes the photoelectric denoising step proposed in~\cite{ying}, but does not apply the calibration steps discussed in~\cite{ying}.   However, our proposed dual-energy results also are uncalibrated, which aids side-by-side comparison. Numerically, the approach solves a constrained least-squares problem separately for each point in the sinogram, and applies  non-negativity constraint to ensure the derived values are physically reasonable.  As proposed in~\cite{ying} and implemented in our YNC code, this constraint is applied by first solving an unconstrained problem, then zeroing out one of the coefficients if negative results are obtained.  The presence of  zeroed regions in the sinogram can produce substantial reconstruction artifacts.  To address this, we applied a simple inpainting algorithm~\footnote{Package `inpaint nans', J. D'Erricco, Mathworks File Exchange} to interpolate across zeroed points in the sinogram.  We note that these artifacts are greater in our system than in systems designed specifically for dual-energy, because the  relatively low energy separation between the two scan energies.

\subsection{Simulation results}

A suitcase phantom was created~\footnote{by Dr. Taly Gilat-Schmidt of Marquette University} and used to generate simulated dual-energy data for the Imatron C300 scanner, assuming the spectra are as shown in Figure~\ref{fig:spectra}.   
 We added Gaussian noise to the Poisson-distributed simulated data to achieve 70 dB signal-to-noise ratio for electronics noise.  The phantom suitcase consists of a plastic outer case surrounding an aluminum block (center of image), a C-shaped neoprene rubber sheet, and a plastic bottle containing water.

\begin{figure}[H]
\centering
\includegraphics[scale=0.31]{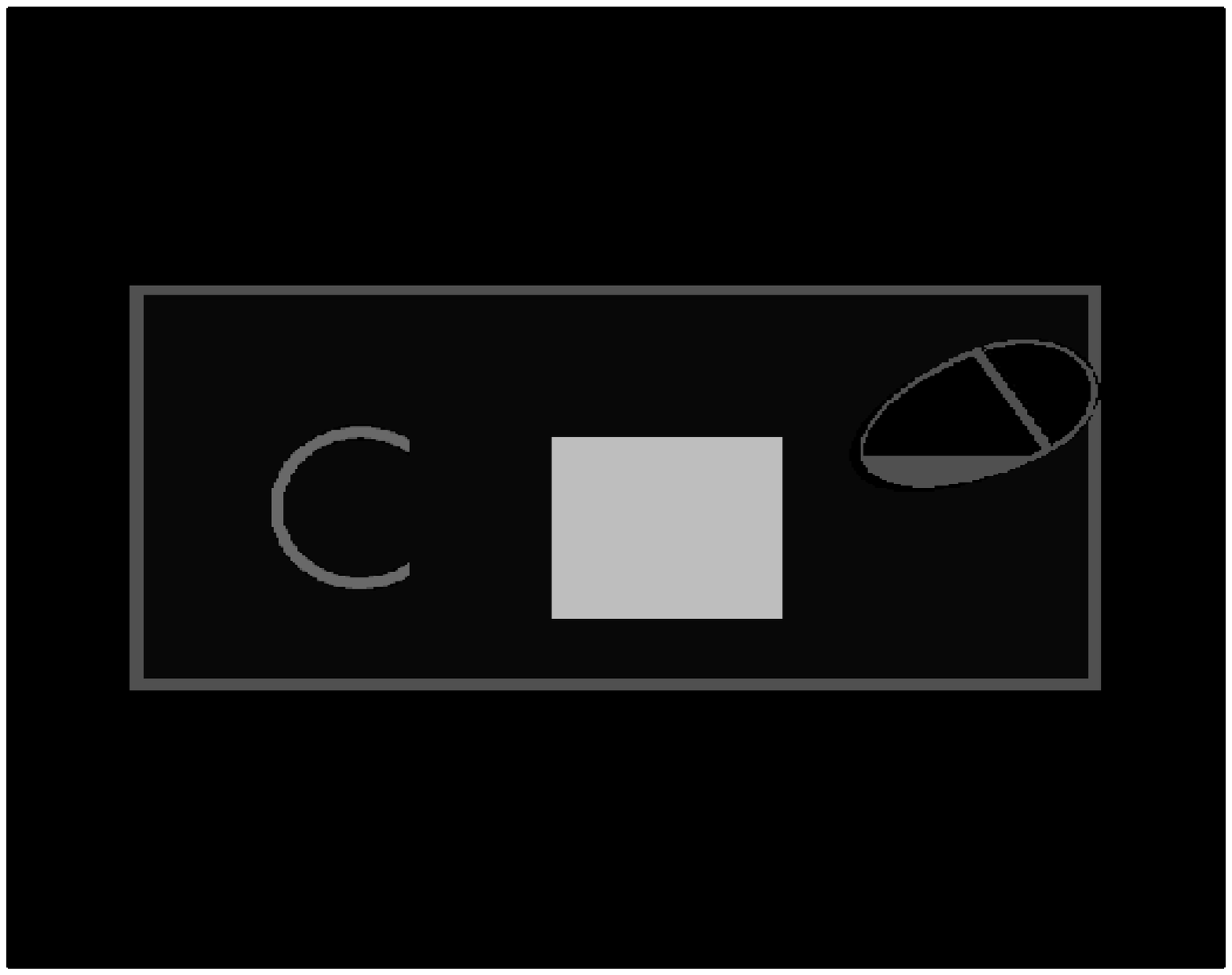}{(a)}
\includegraphics[scale=0.31]{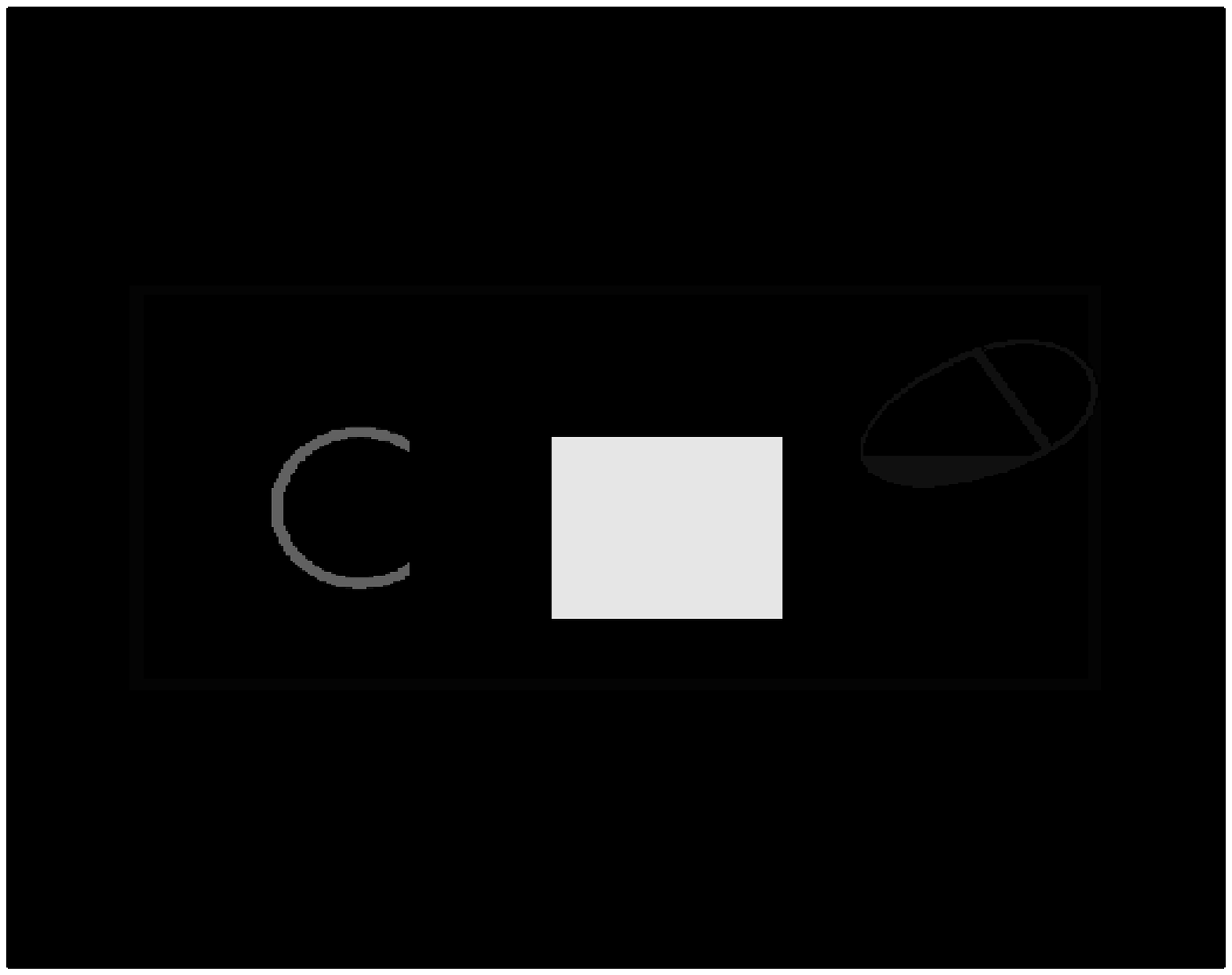}{(b)}
\includegraphics[scale=0.31]{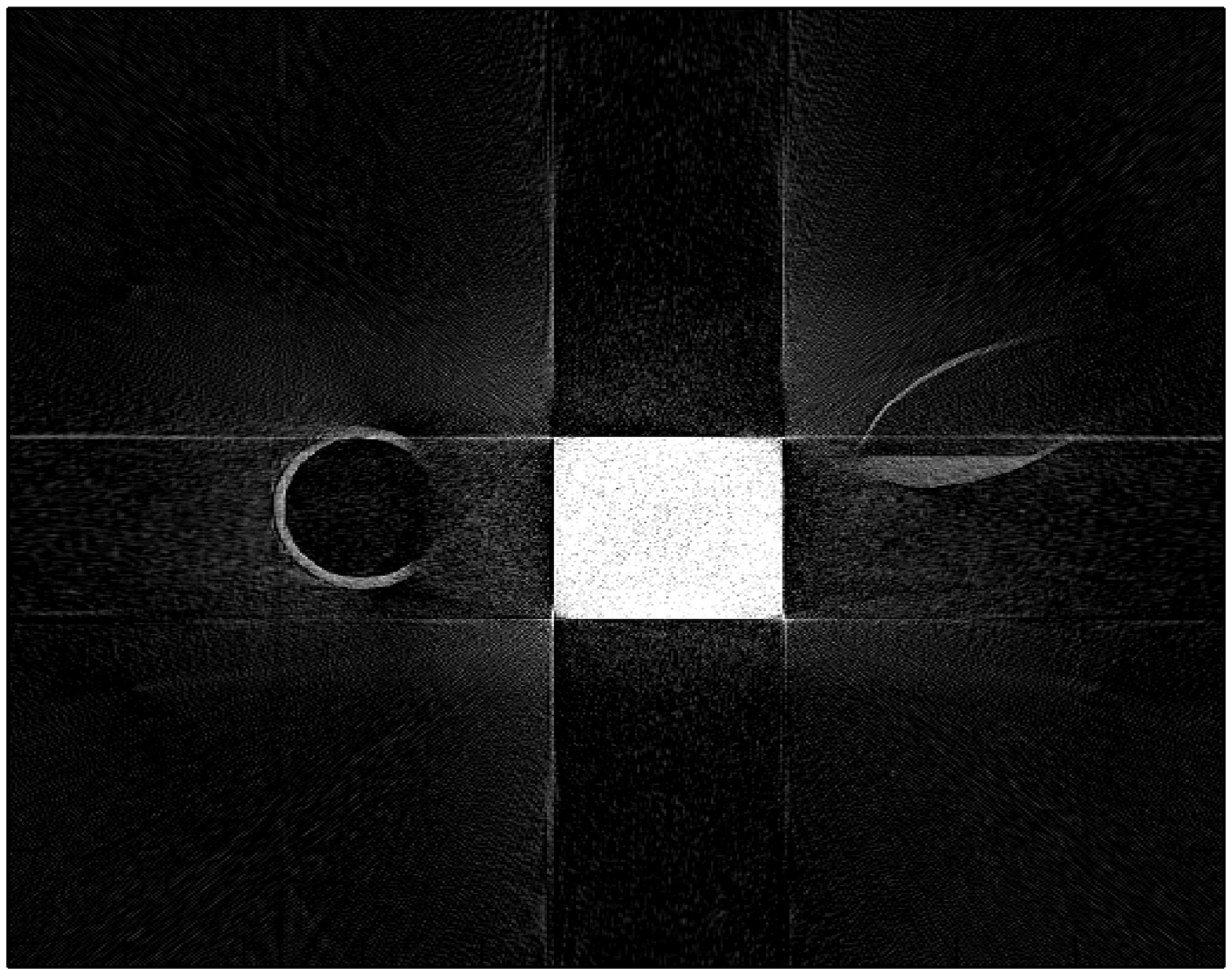}{(c)}
\includegraphics[scale=0.31]{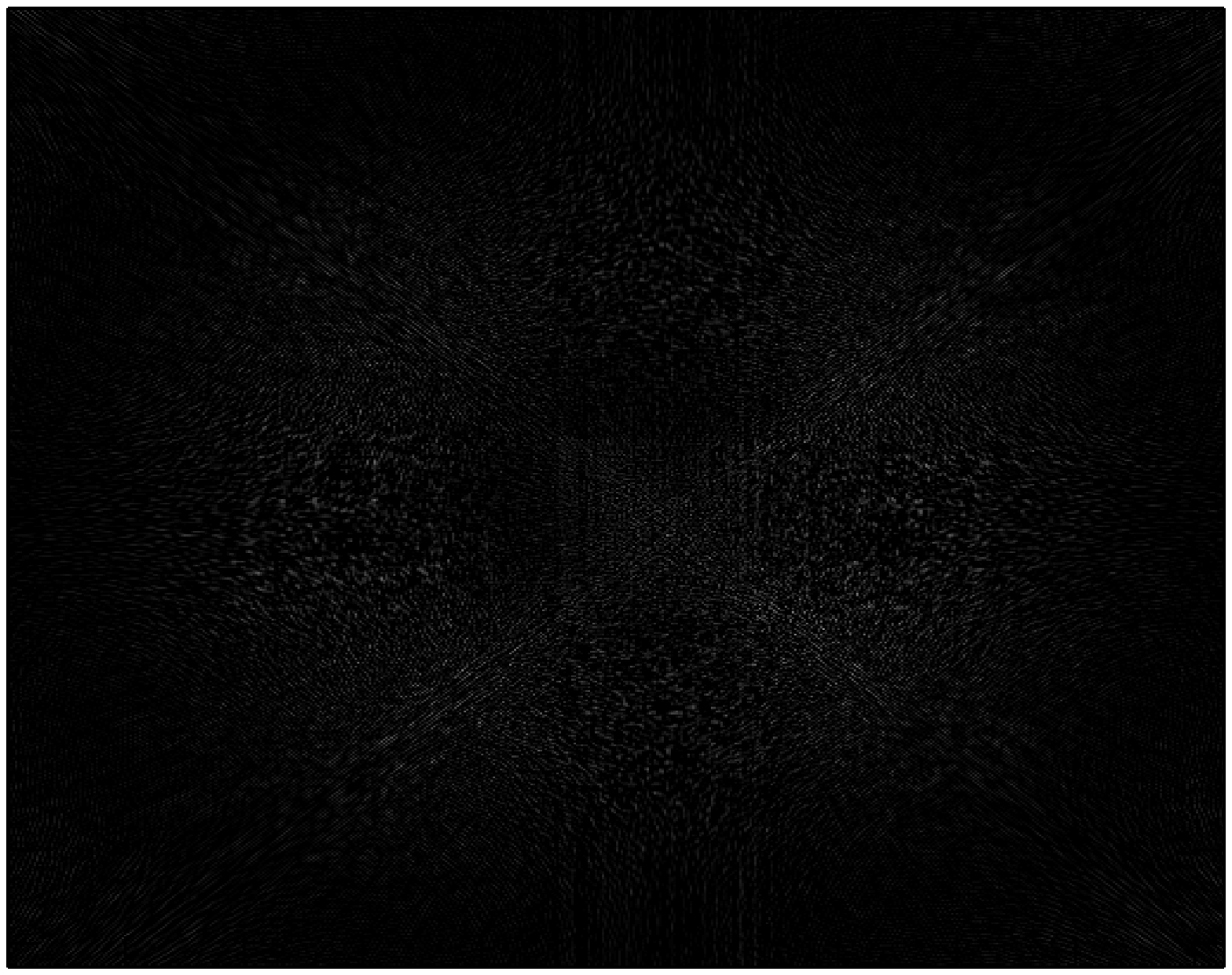}{(d)}
\includegraphics[scale=0.31]{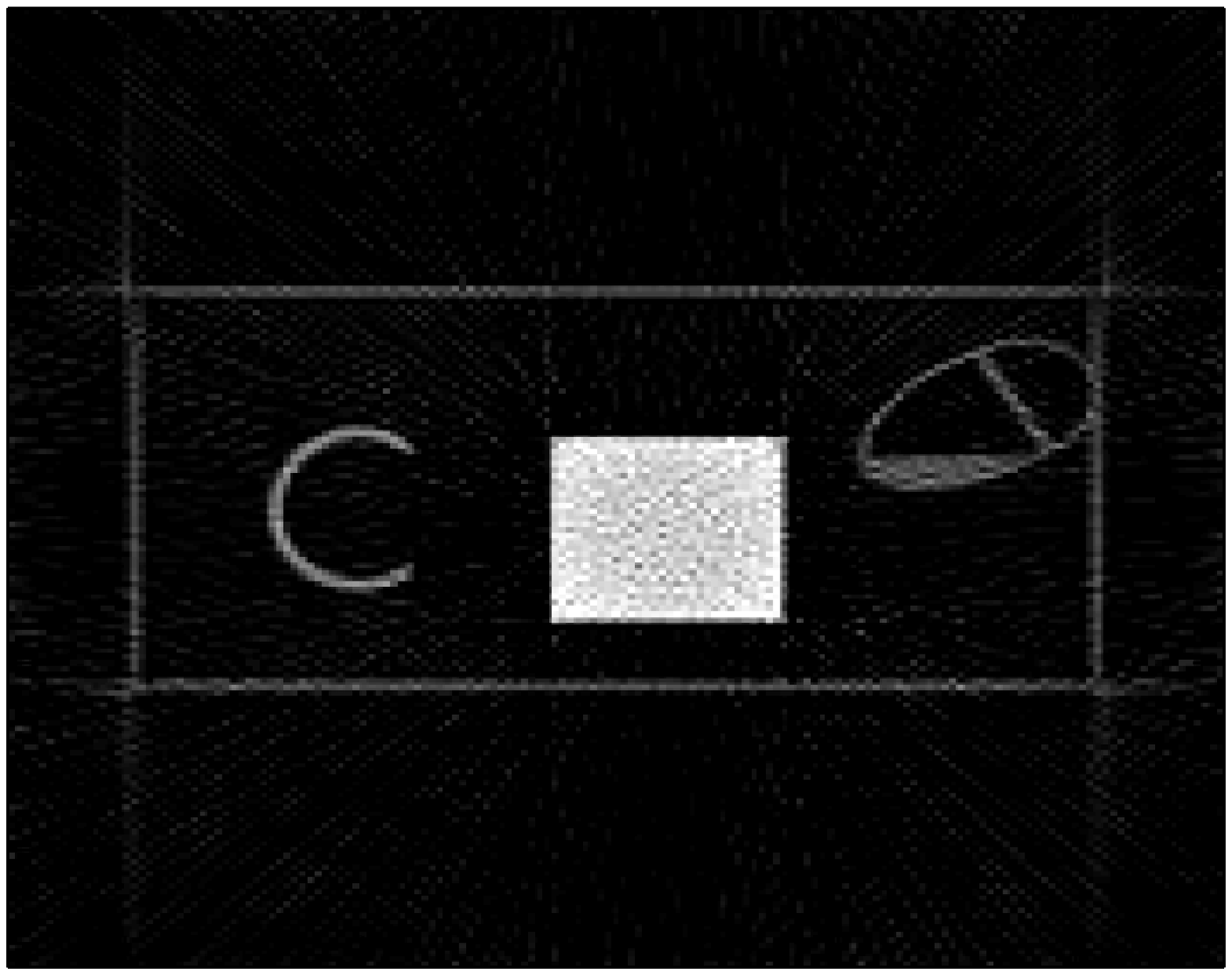}{(e)}
\includegraphics[scale=0.31]{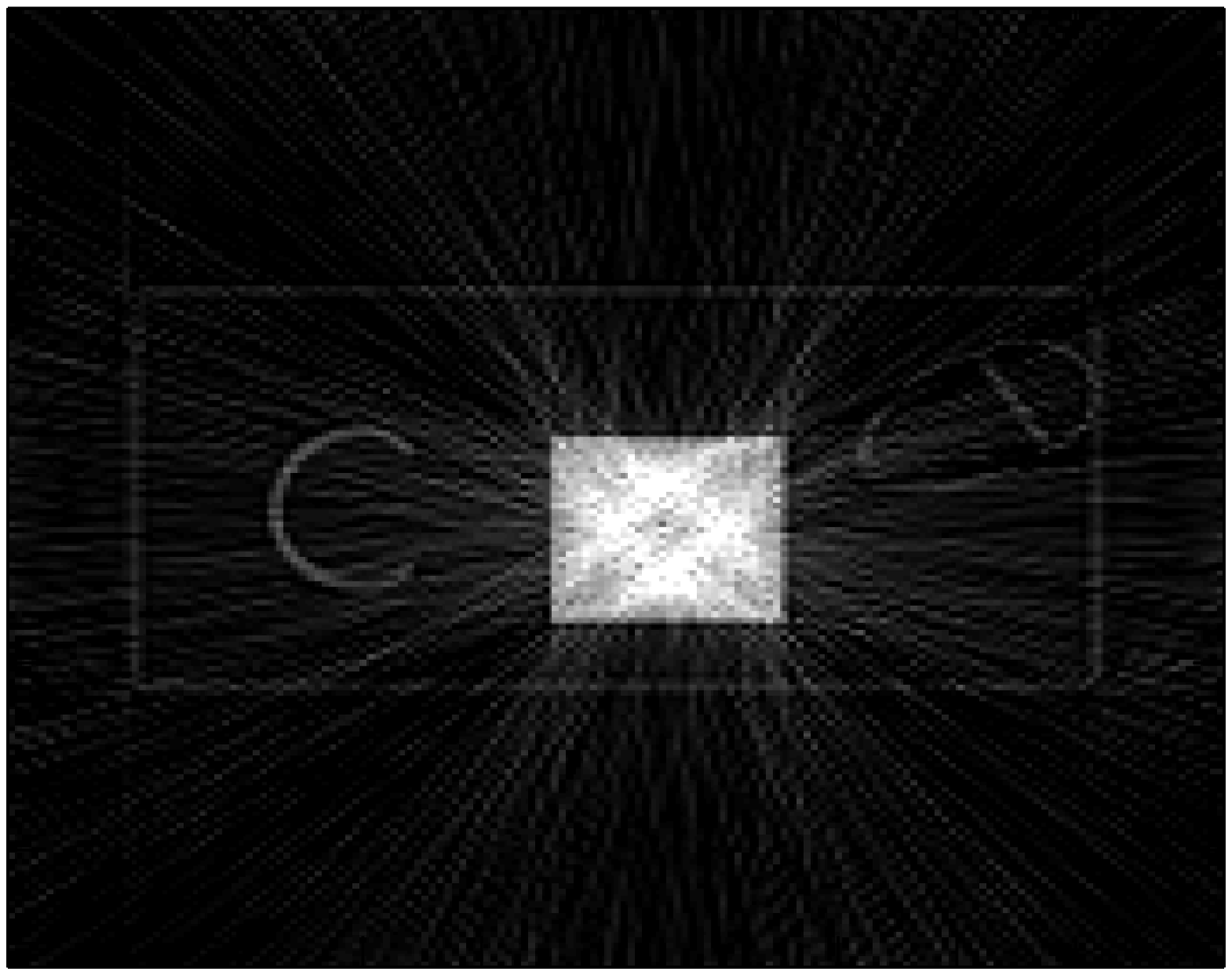}{(f)}
\includegraphics[scale=0.31]{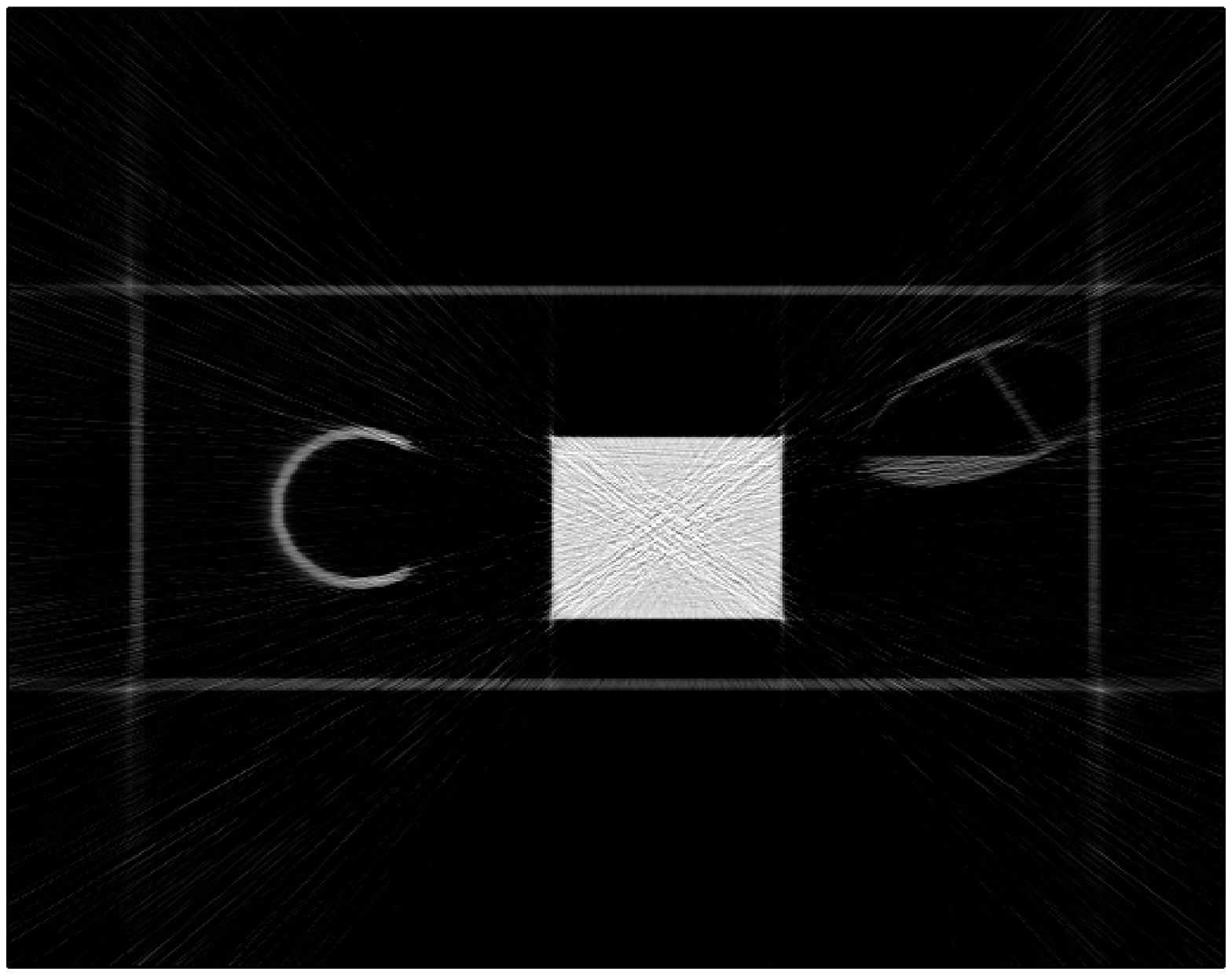}{(g)}
\includegraphics[scale=0.31]{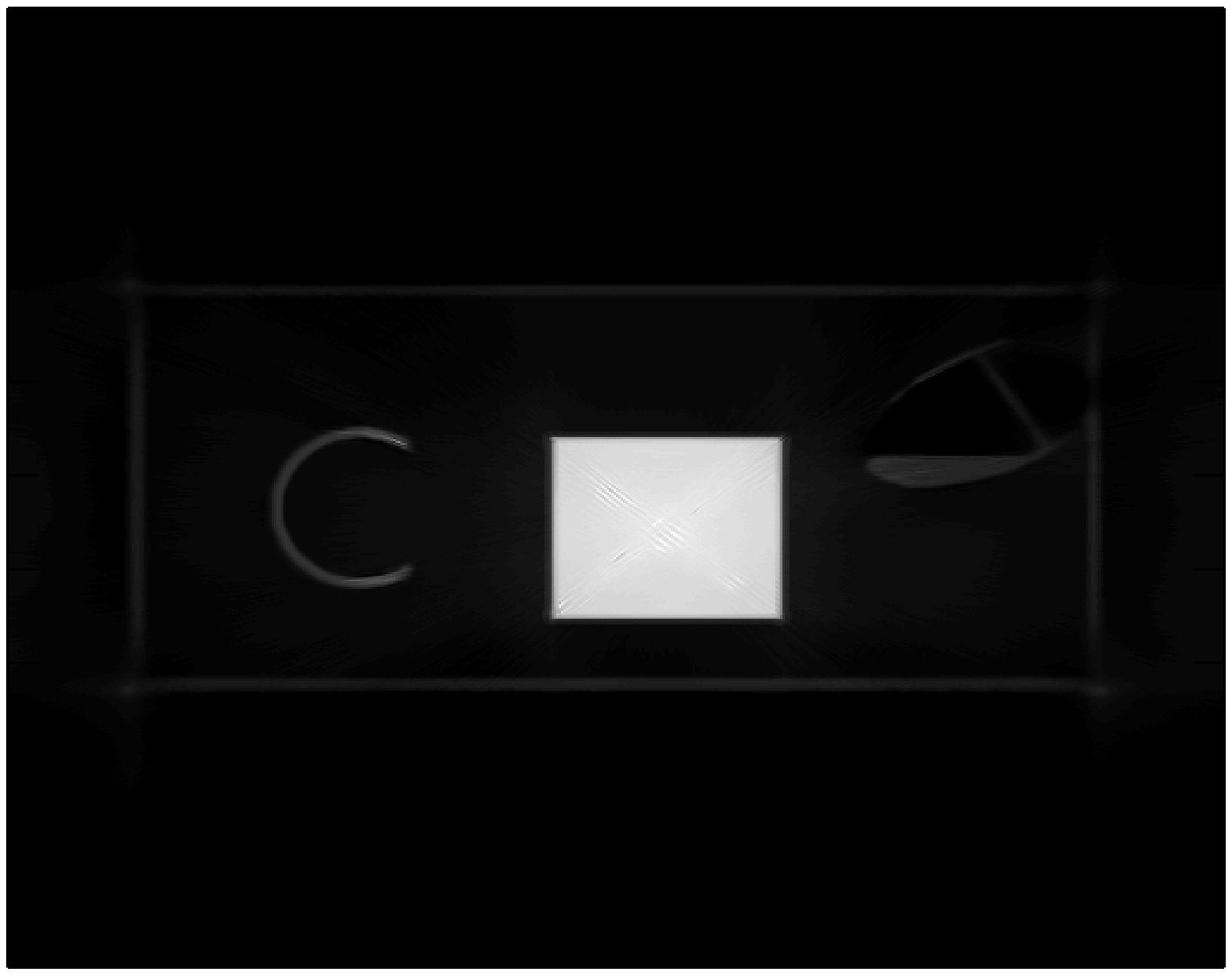}{(h)}
\includegraphics[scale=0.31]{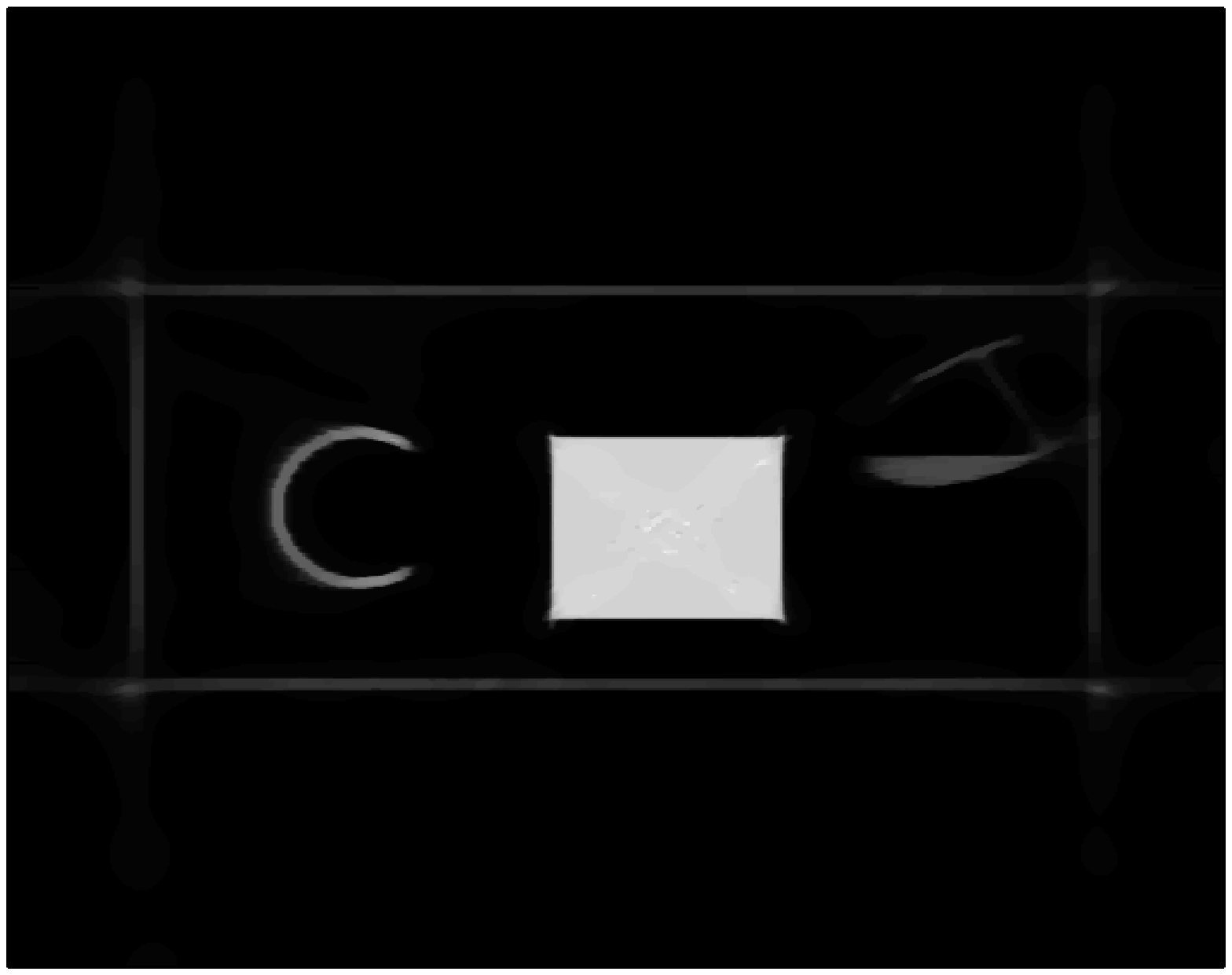}{(i)}
\includegraphics[scale=0.31]{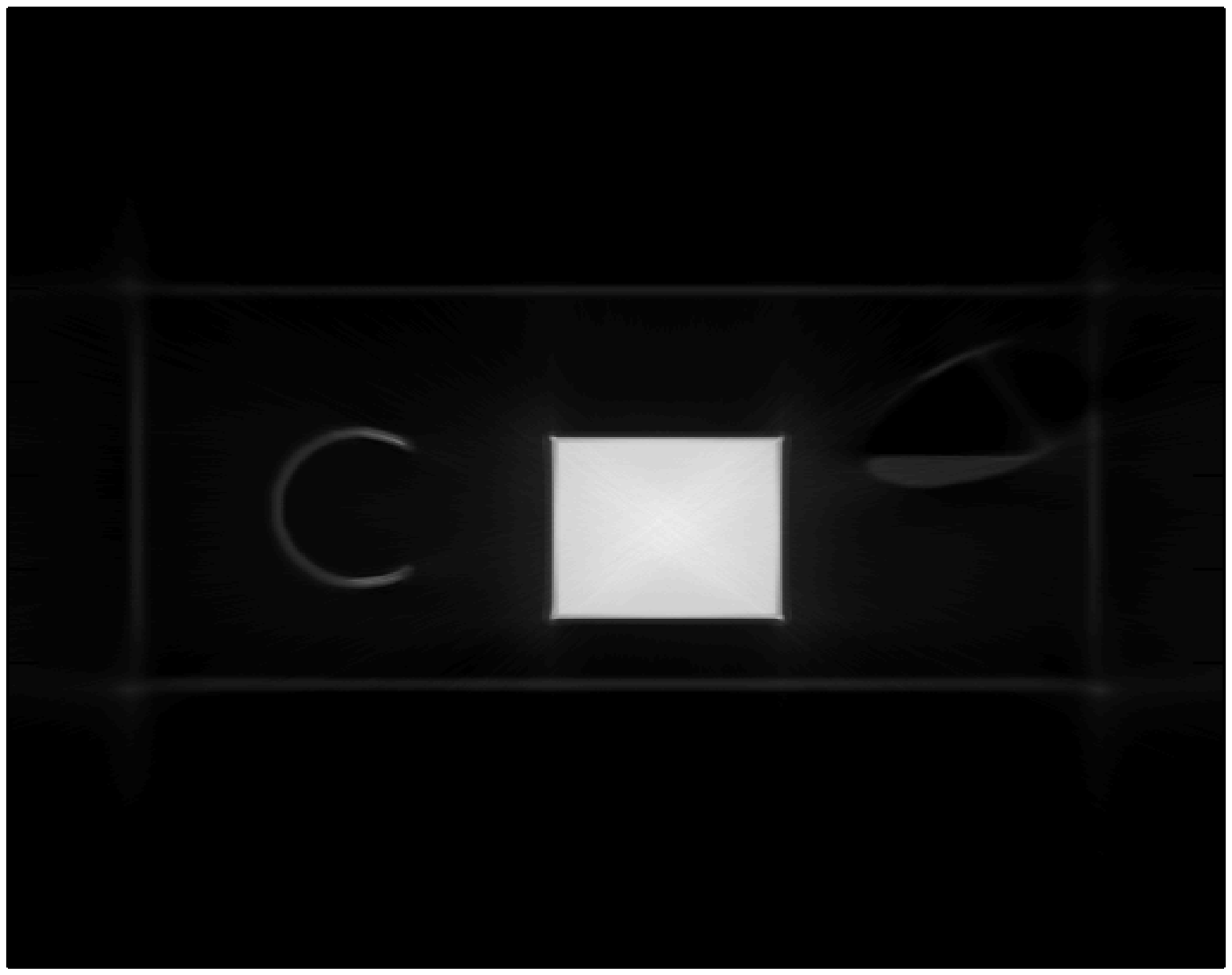}{(j)}
\caption{Compton images (left) and photoelectric images (right) for  suitcase phantom.  Top row (a,b) is ground truth; 2nd row is the YNC result; 3rd row is Levenberg-Marquart solution, using the edge-based regularization of~\cite{Semerci2012}; 4th row is the proposed ADMM solver with patch-based regularization of photoelectric; and bottom row is the proposed method but additionally applying Total Variation regularization to the Compton image.}.
\label{fig:simSuitcase}
\end{figure}

Figure~\ref{fig:simSuitcase} show Compton (left) and photoelectric (right) coefficient images for a series of reconstruction approaches.  In each case the Compton image is plotted on the range $0-0.7 
 cm^{-1}$, while the photoelectric image is plotted for the range $0-8e^4 KeV cm^{-1}$. 
 The upper row in Fig.~\ref{fig:simSuitcase} shows the ground truth for the scenario, while the second row shows YNC results.  While the Compton image is partially recovered by YNC, there is essentially no structure in the recovered photoelectric image.  

As discussed, our group previously investigated regularizers based on edge correlation between Compton and photoelectric images~\cite{Semerci2012}.  The third row of Fig.~\ref{fig:simSuitcase} shows the cyclic descent Levenberg-Marquart approach described in \cite{Semerci2012} using the edge correlation regularizer.   This approach gives a greatly improved photoelectric image, but there is noticeable noise in the image and the outline of the object to the upper right (simulated water bottle) is obscured.  Using our proposed patch-based regularizer in the ADMM framework we obtain the fourth row in Fig.~\ref{fig:simSuitcase}.  The NLM regularizer leads to a much smoother photoelectric estimate  while the Compton image is little  changed.  In this result we apply non-negativity constraints to the Compton image but do not apply the Total Variation penalty.  One important difference between the two approaches to regularizing the photoelectric image is that, because the edge-based regularizer encourages photoelectric solutions with similar edges to the Compton image, any streak artifacts in the Compton image will tend to carry over to the photoelectric image.  The NLM regularizer, by contrast, encourages averaging during the photoelectric reconstruction of patches that are similar in the Compton image, but is less likely to map streak artifacts from Compton to photoelectric.

Finally, the last row in  Fig.~\ref{fig:simSuitcase} shows our proposed method which uses the ADMM solution and includes TV and non-negativity constraints on the Compton image as well as patch-based regularization of the photoelectric image.  As can be seen, the TV penalty leads to a more homogeneous Compton image for the aluminum block.

Because ground truth is known in this simulation example, the reconstructions methods can be compared using metrics such as peak signal-to-noise ratio (PSNR) and the structural similarity index (SSIM)~\cite{SSIM}.  PSNR is defined by comparing image $\im{I}$ to reference image $\im{I}_{ref}$ as
\begin{equation}
PSNR = 10 \log_{10} \frac{L^2}{1/N\sum_k (I(k)-I_{ref}(k))^2}
\end{equation}
where $j$ sums over the $N$ pixels in the image, and $L$ is the expected maximum value in the image.   Here $L$ is taken to be $0.7$ for Compton images and $1.2e5$ for photoelectric images, values which are chosen as they exceed the maximum values in any of the estimated images.  Table~\ref{tab:metricsReg} shows the PSNR AND SSIM metrics corresponding to Figs.~\ref{fig:simSuitcase}.  The benefit of patch-based regularization is clear in photoelectric image results.  Both outperform the YNC approach.  The two ADMM solutions are very close and outperform the other options. 

\begin{table}
\begin{center}
\begin{tabular}{|l|c|c|}
\hline
\bf{Regularization type} & \bf{Compton image} & \bf{Photoelectric  image}\\\hline
YNC~\cite{ying}  & 21.8 (0.27)  &  18.9 (0.10)  \\ 
Edge-based, Levenberg-Marquart \cite{Semerci2012}  & 25.7 (0.28)  &  25.8 (0.27)  \\
ADMM, NLM regularization  & 29.5 (0.69)  &  32.5 (0.62)  \\ 
Proposed (ADMM with NLM and TV)  & 30.9 (0.74)  &  31.8 (0.65)  \\ 
\hline
\end{tabular}
\caption{Effect of regularization type on image quality, for results plotted in Figs.~\ref{fig:simSuitcase}.  Metrics shown are PSNR in dB, and SSIM in parentheses.}
\end{center}
\label{tab:metricsReg}
\end{table}

\subsection{Experimental data}

A series of test suitcases were assembled by subject matter experts and scanned on the Imatron scanner~\footnote{These subject matter experts were assigned by the ALERT center at Northeastern University, which directed this work.}.  The various bags were packed to contain representative objects from the stream of commerce, including liquids, bottles containing beads (providing examples of textured objects), and rubber sheets.  The experts also identified a set of homogeneous objects in the scans (water, doped water, and rubber sheets) which are used in multiple scans and have identical material properties.  By comparing the estimates made for these materials, quantitative comparison of various methods can be made, as reconstructions ideally should produce material parameter estimates for these objects which are homogeneous within a scan and repeatable between scans.

\subsubsection{Individual slices}

\begin{figure}[H]
\centering
\includegraphics[scale=0.3]{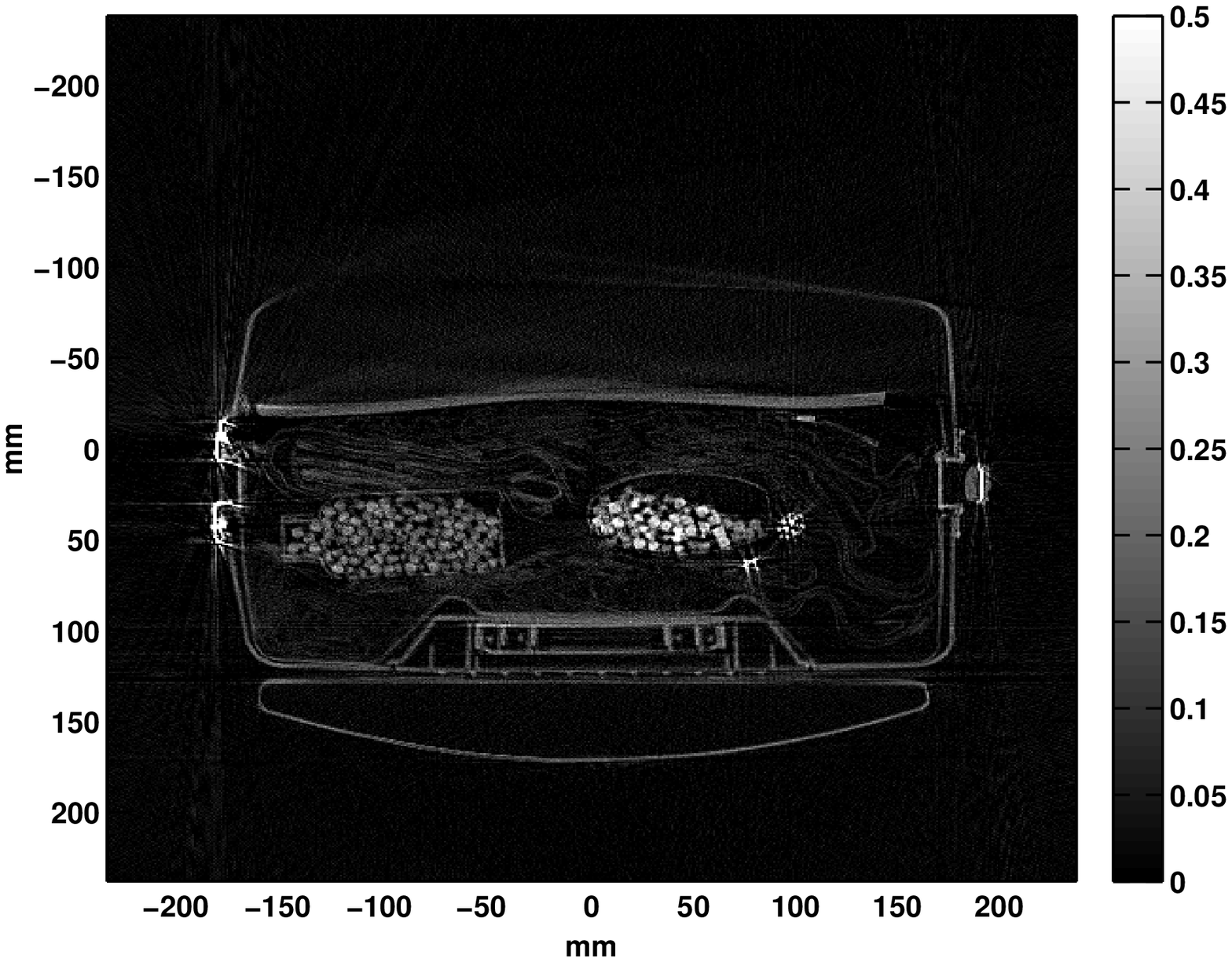}{(a)}
\includegraphics[scale=0.3]{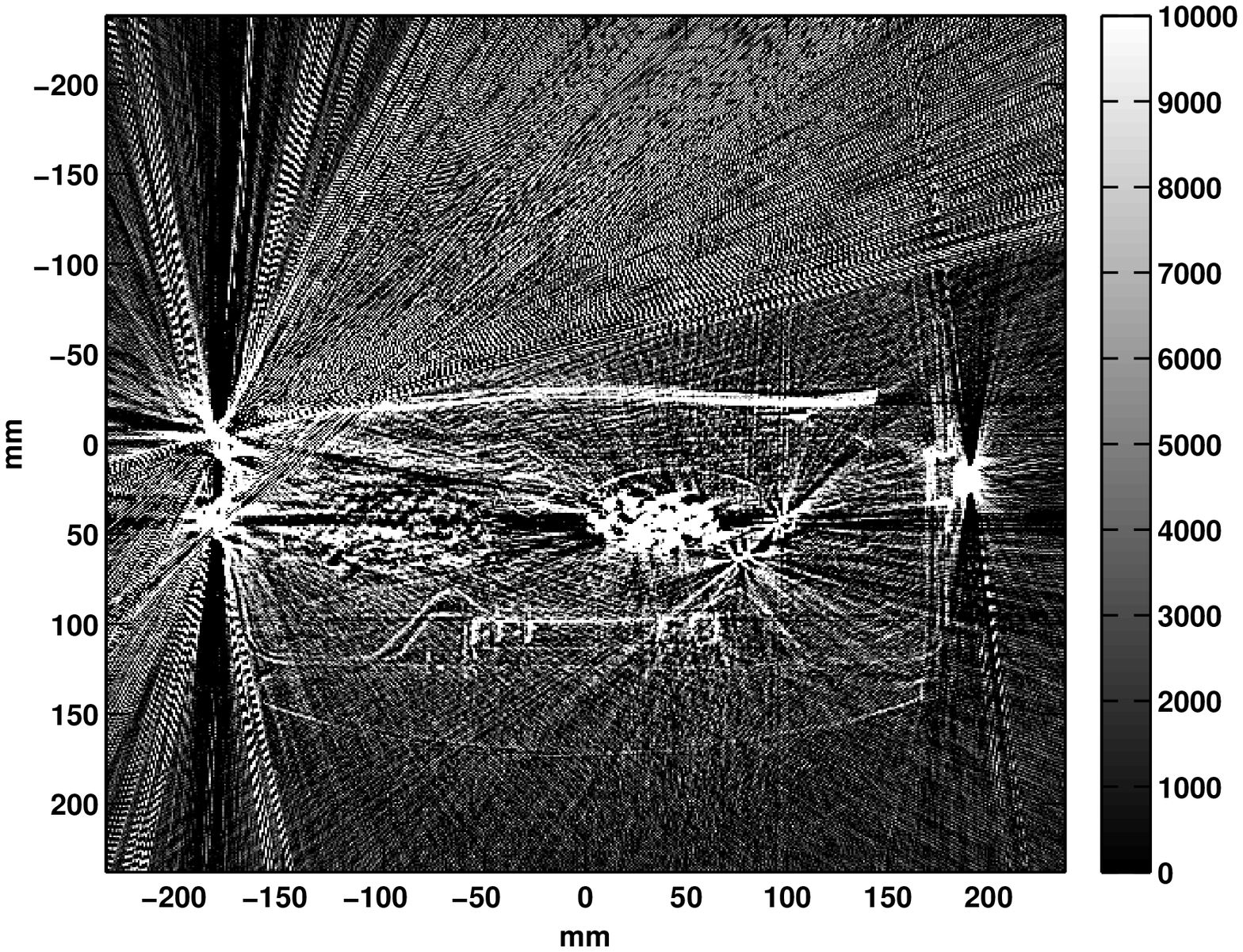}{(b)}
\includegraphics[scale=0.3]{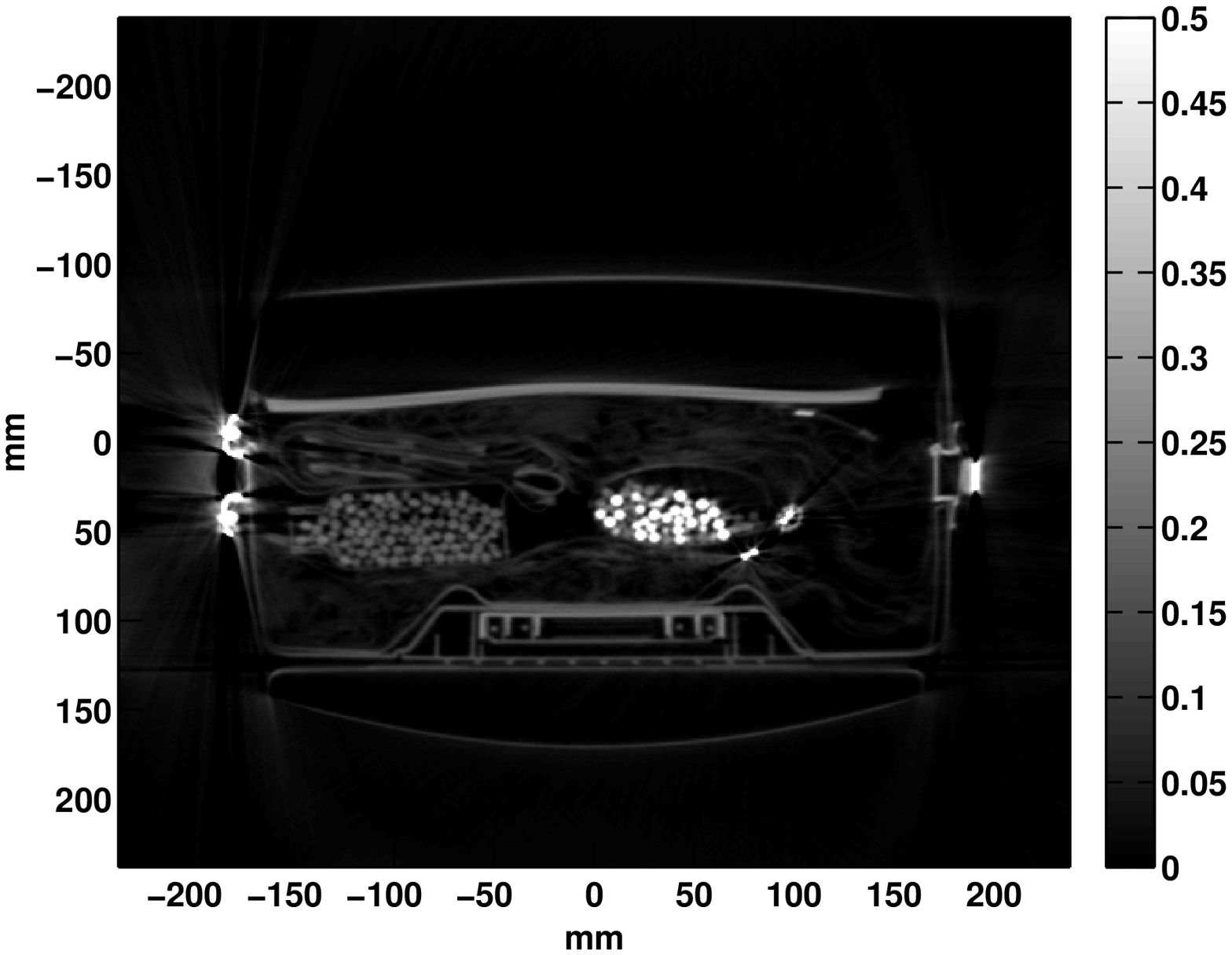}{(c)}
\includegraphics[scale=0.3]{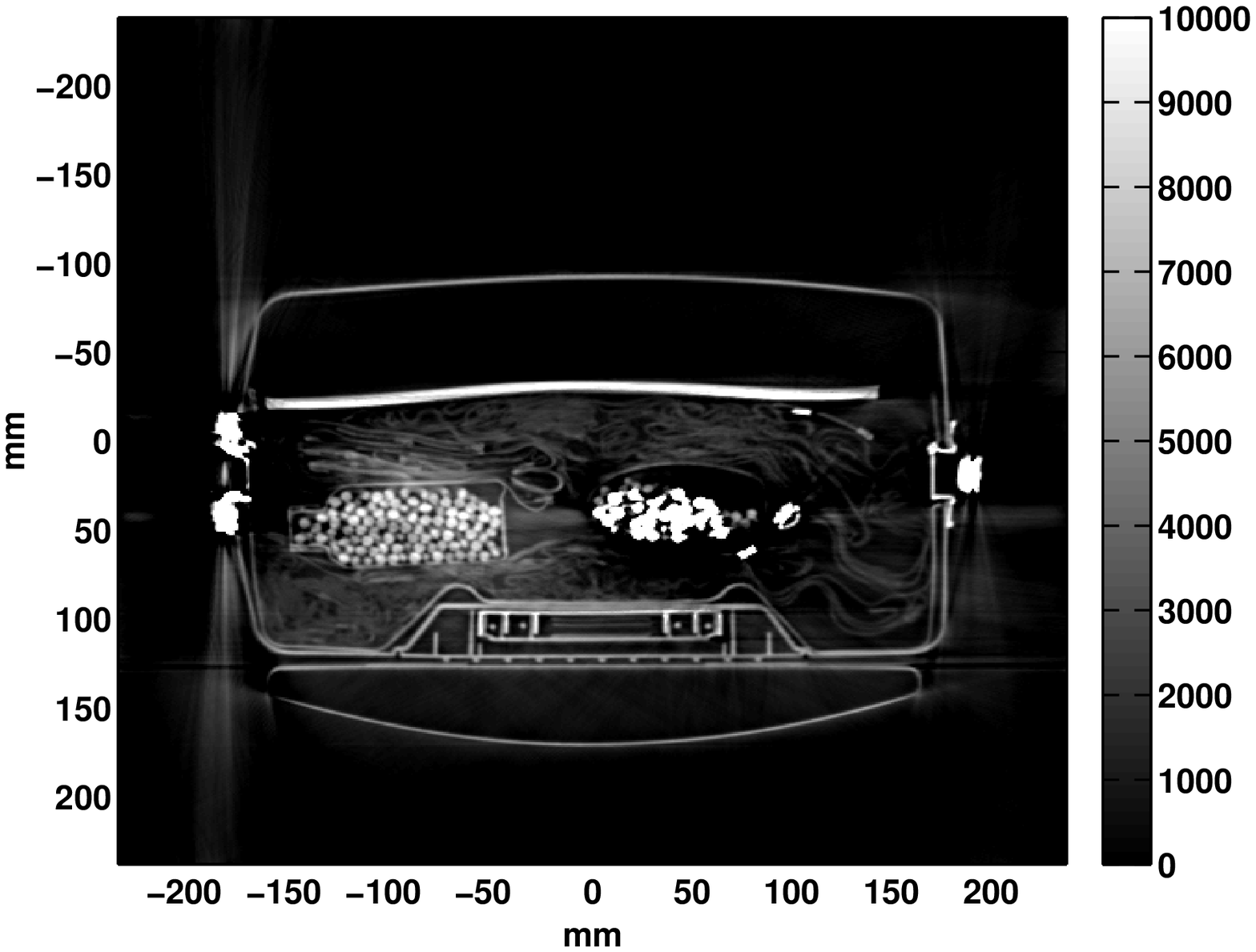}{(d)}
\includegraphics[scale=0.3]{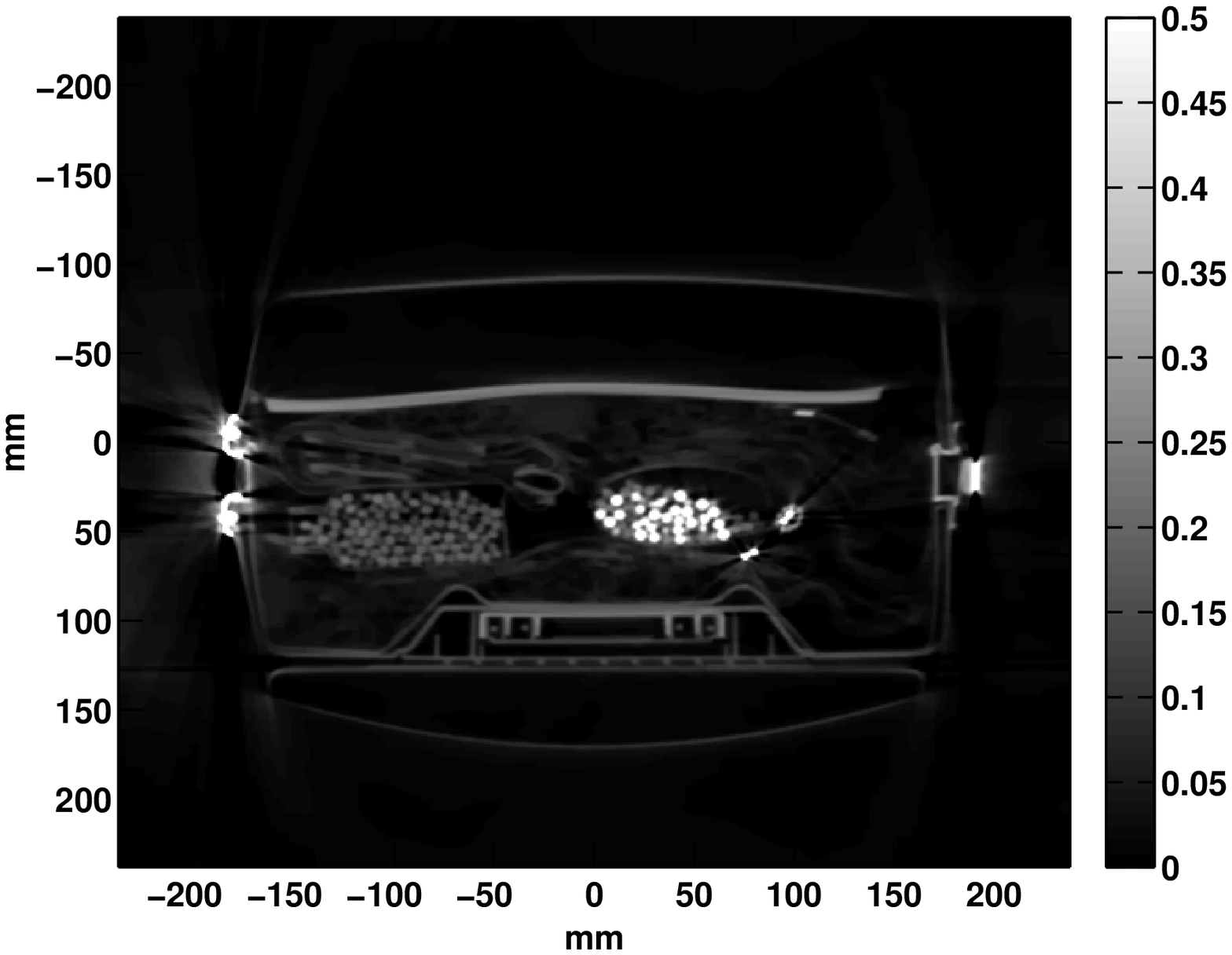}{(e)}
\includegraphics[scale=0.3]{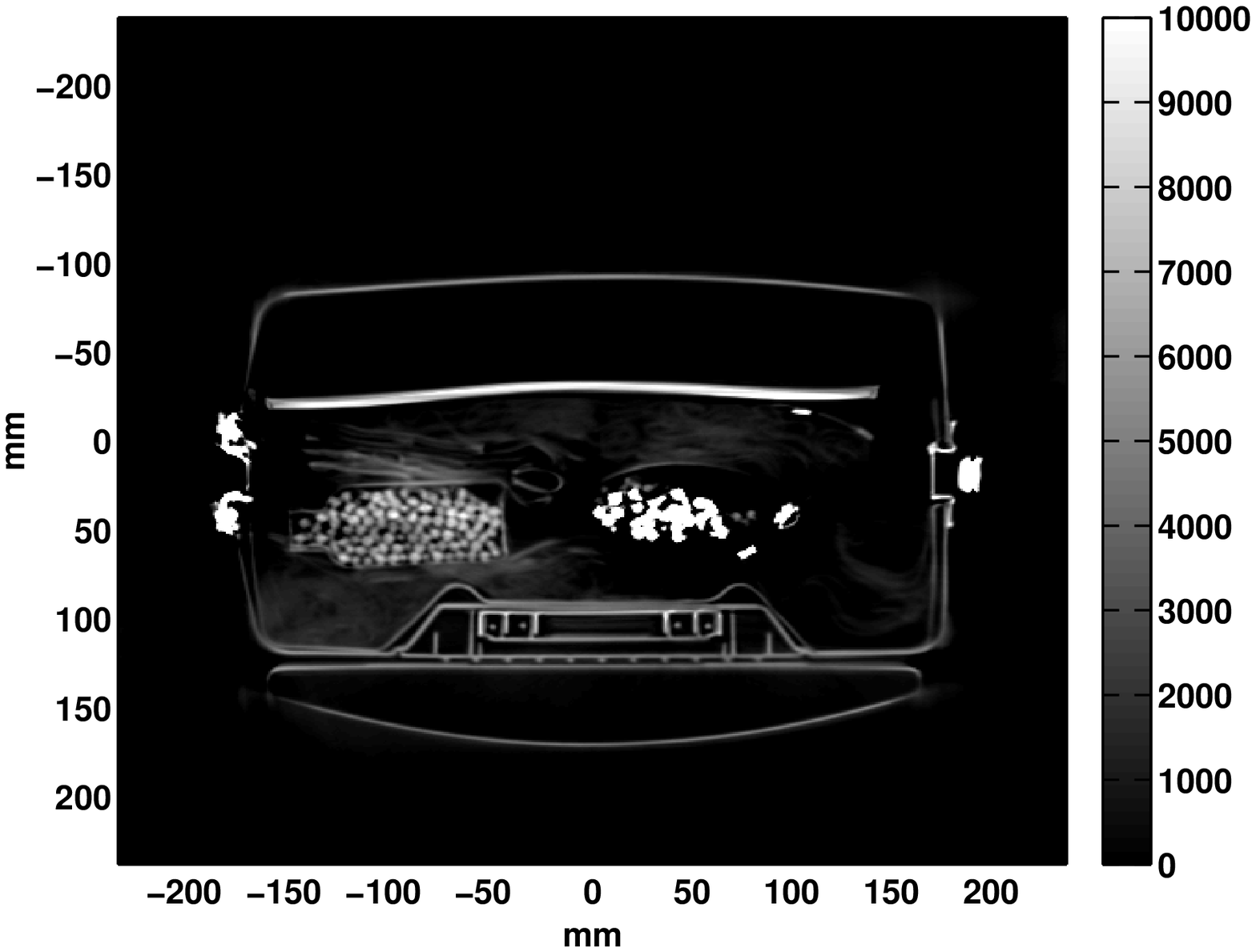}{(f)}
\caption{Packed suitcase with rubber sheet, clothes, and bottle with beads.  Compton images are on left, photoelectric on right.  Top row (a,b) are legacy; 2nd row (c,d) are iterative results but with no regularization; 3rd row (d,e) are iterative results with proposed regularization.}.
\label{fig:MC195}
\end{figure}

\begin{figure}[H]
\centering
\includegraphics[scale=.3]{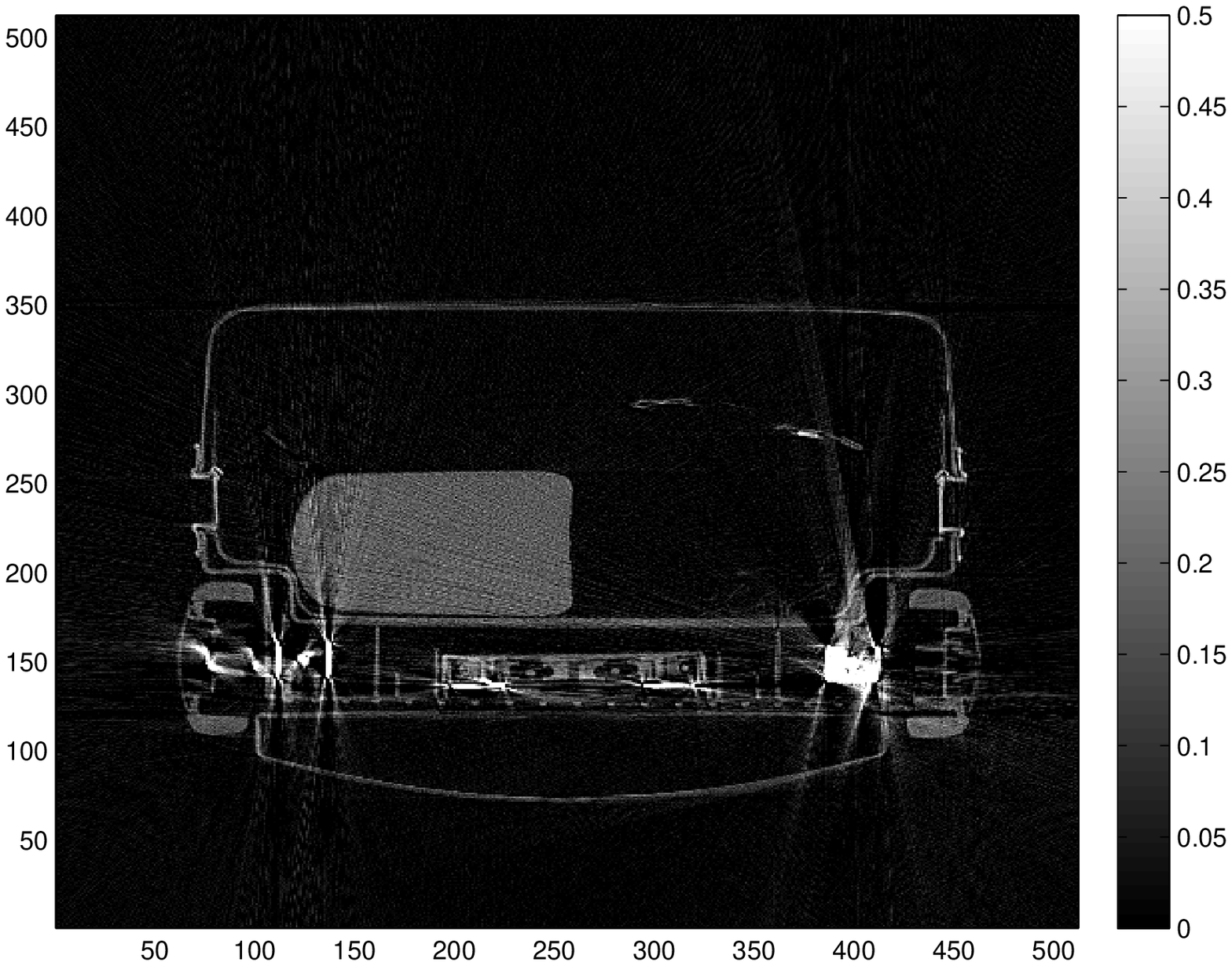}{(a)}
\includegraphics[scale=.3]{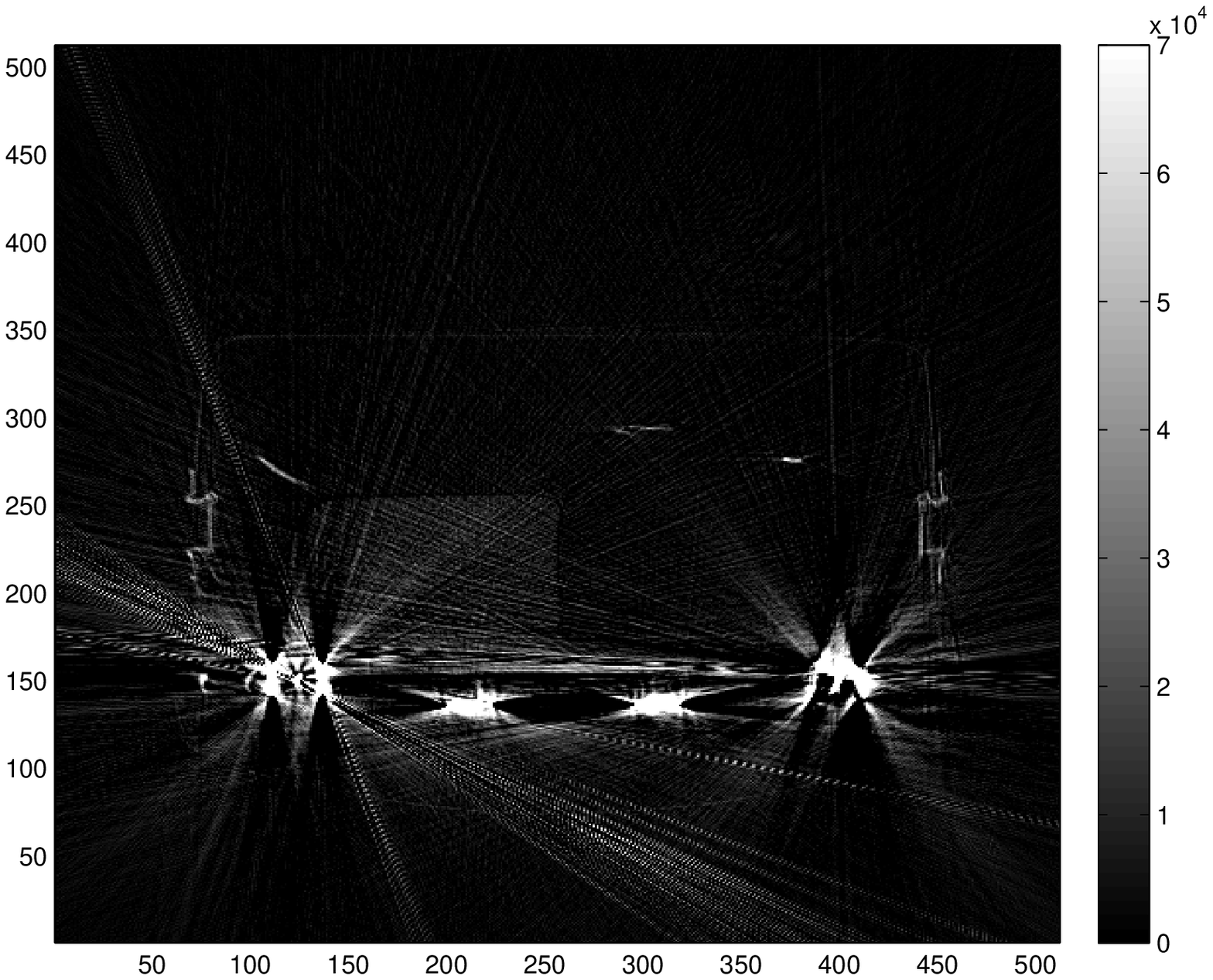}{(b)}
\includegraphics[scale=.3]{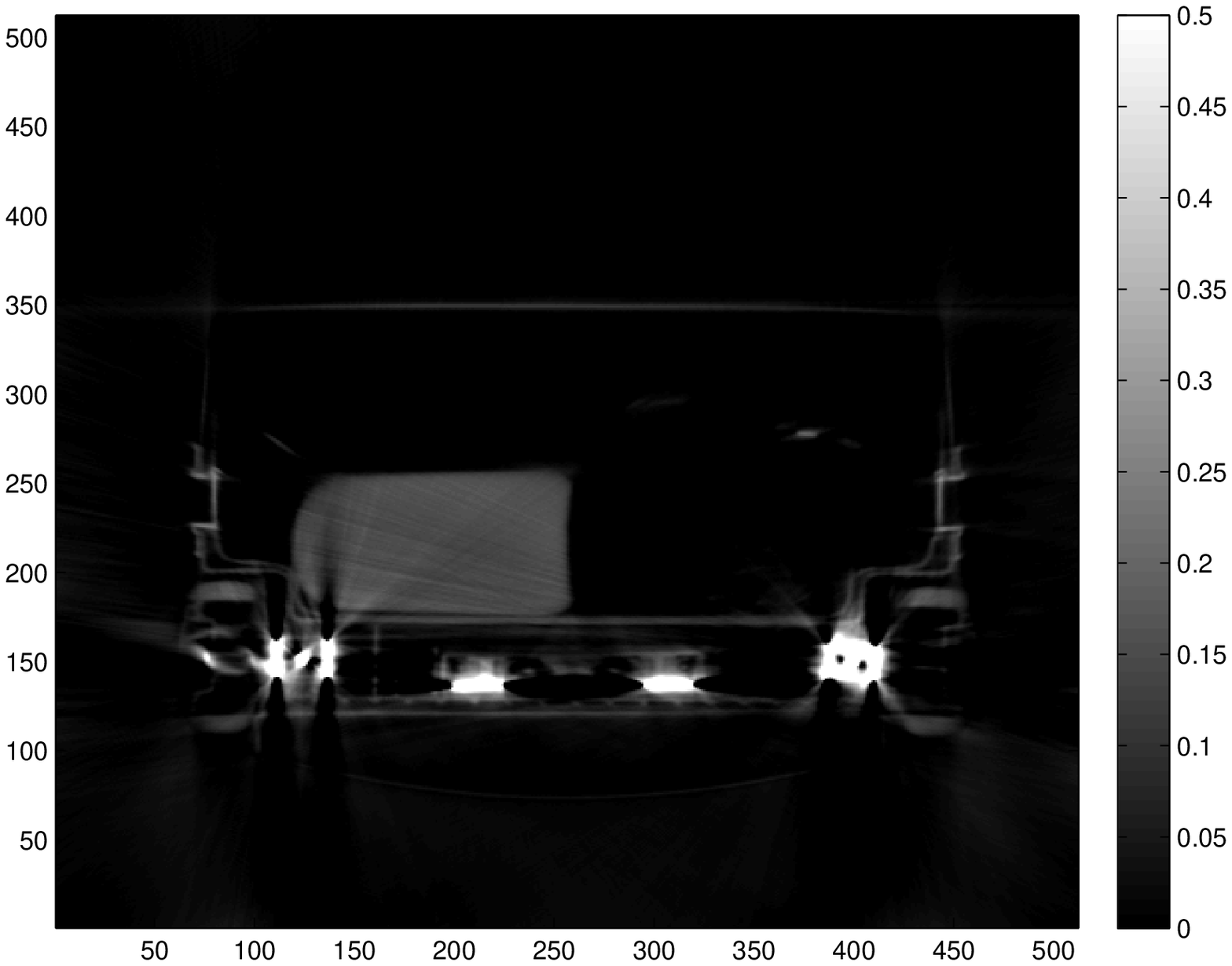}{(c)}
\includegraphics[scale=.3]{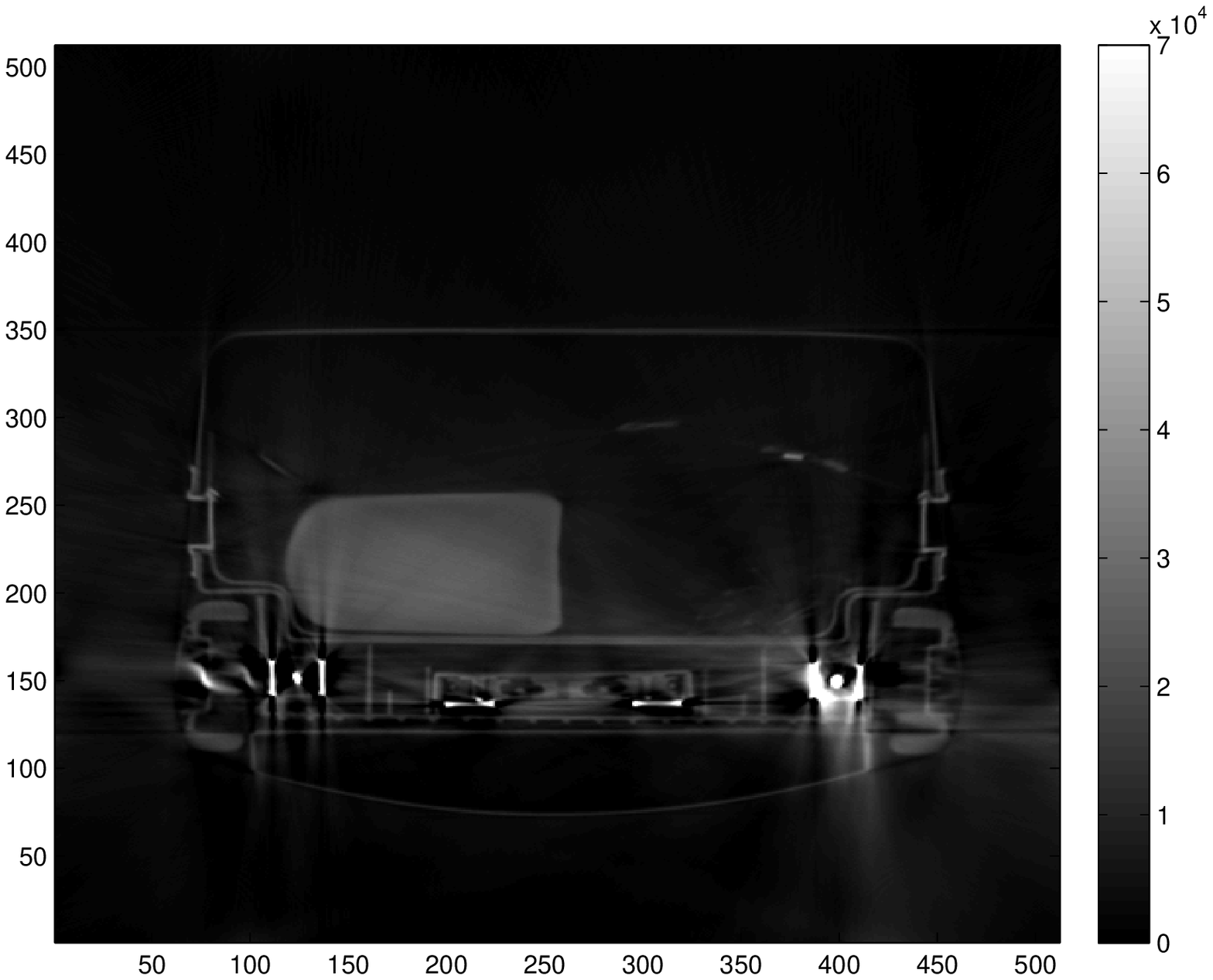}{(d)}
\includegraphics[scale=.3]{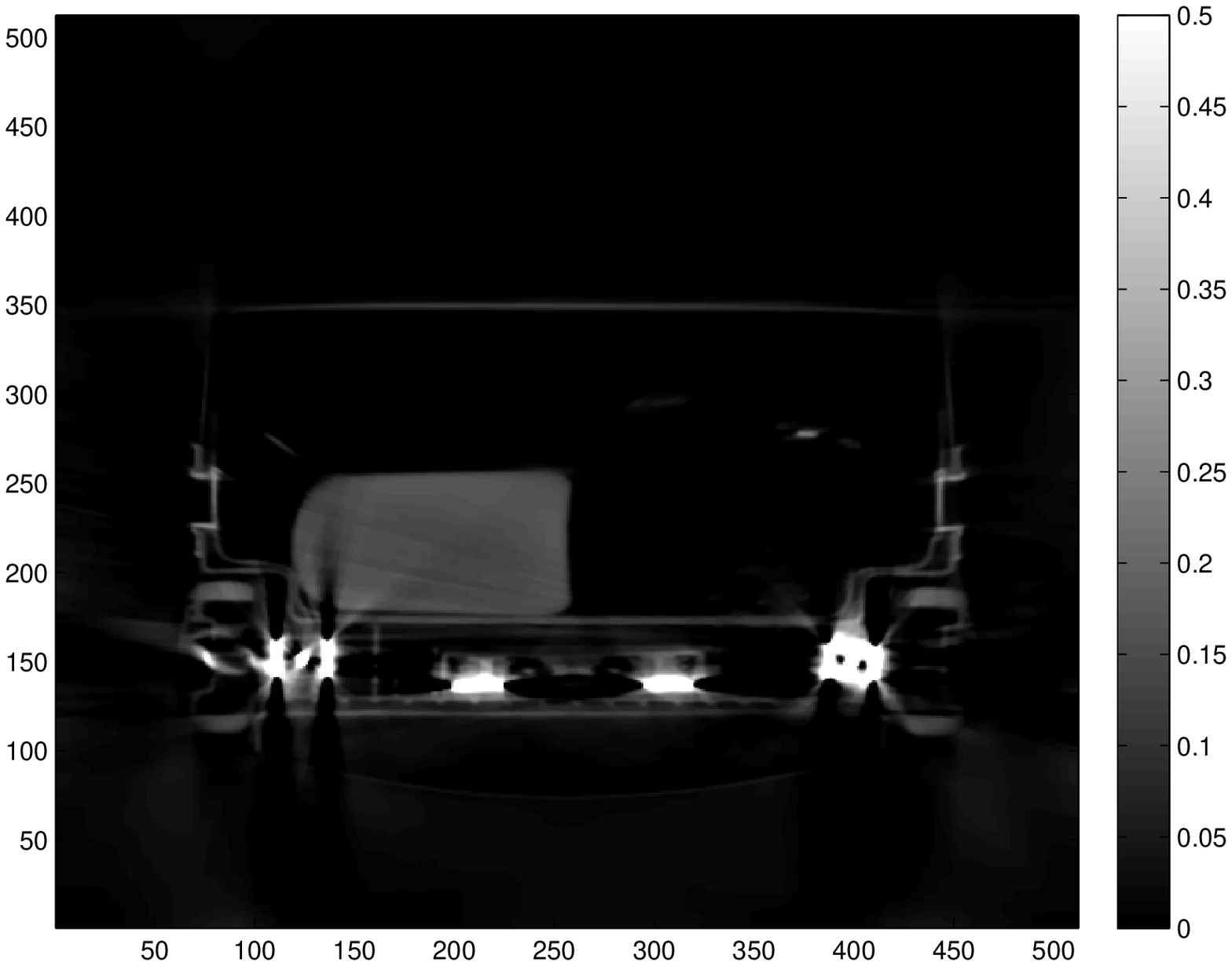}{(e)}
\includegraphics[scale=.3]{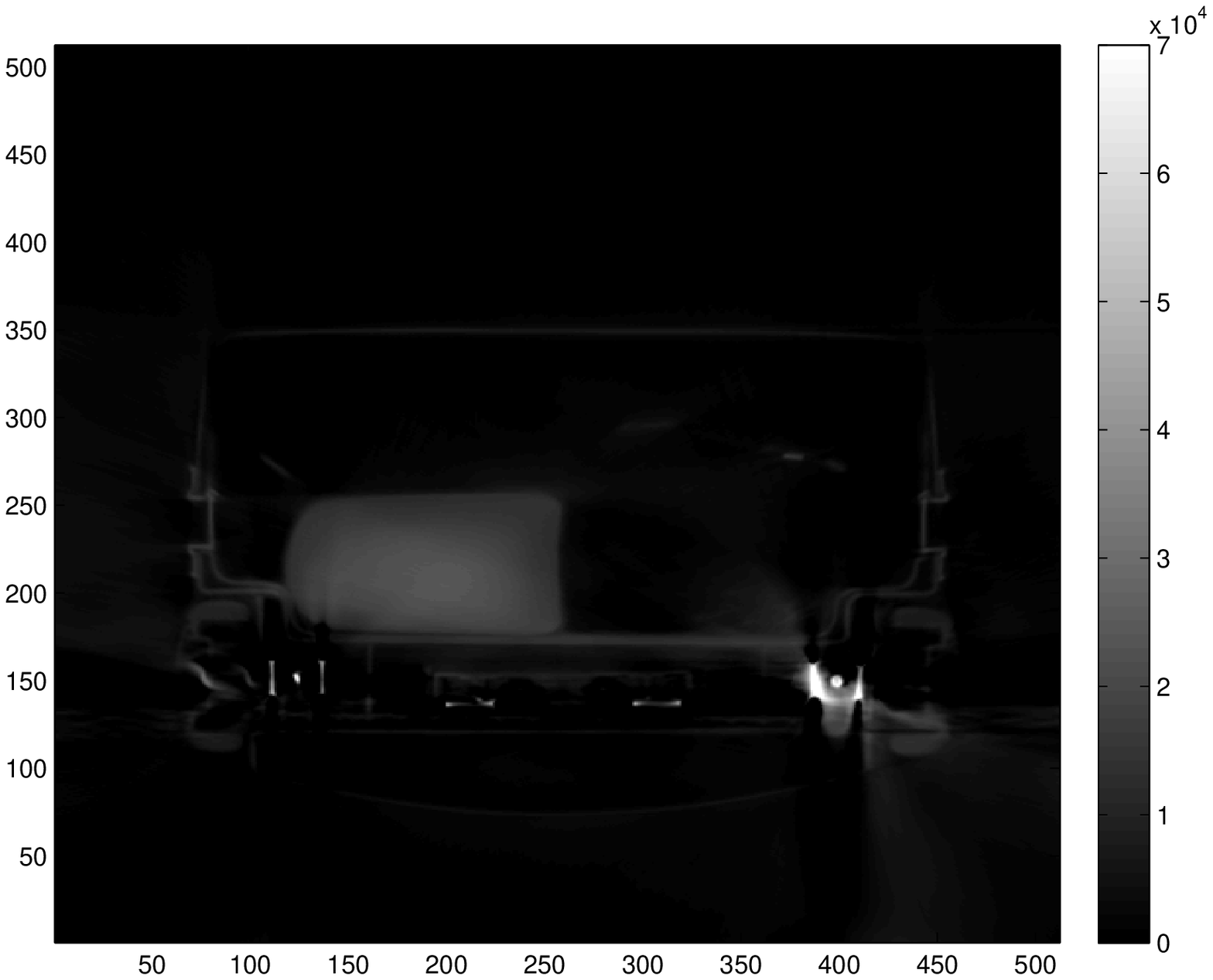}{(f)}
\caption{Lightly packed suitcase with large water bottle (lower left).  Compton images are on left, photoelectric on right.  Top row (a,b) are legacy; 2nd row (c,d) are iterative results but with no regularization; 3rd row (d,e) are iterative results with proposed regularization.} 
\label{fig:mc38}
\end{figure}

\begin{figure}[H]
\centering
\includegraphics[scale=.3]{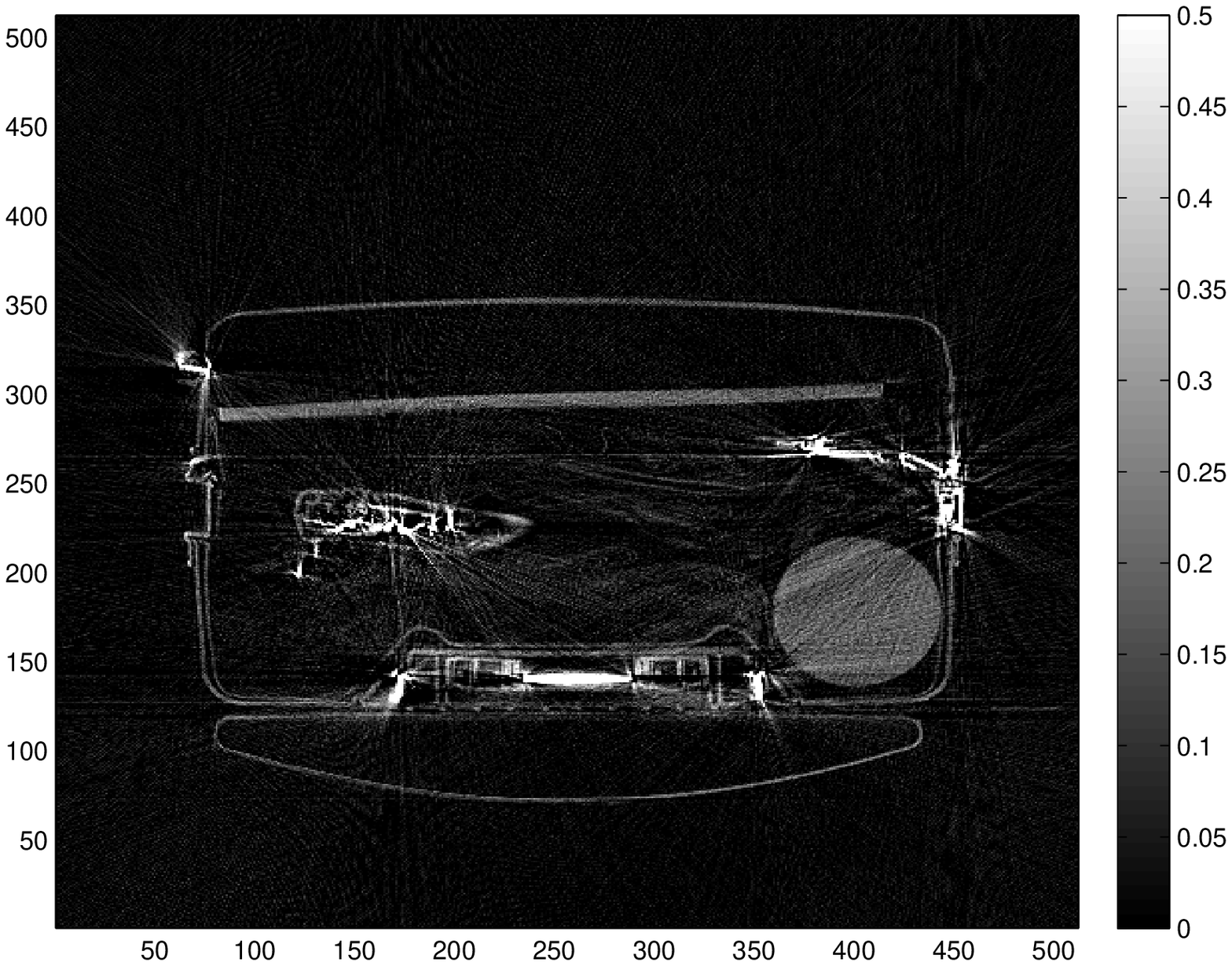}{(a)}
\includegraphics[scale=.3]{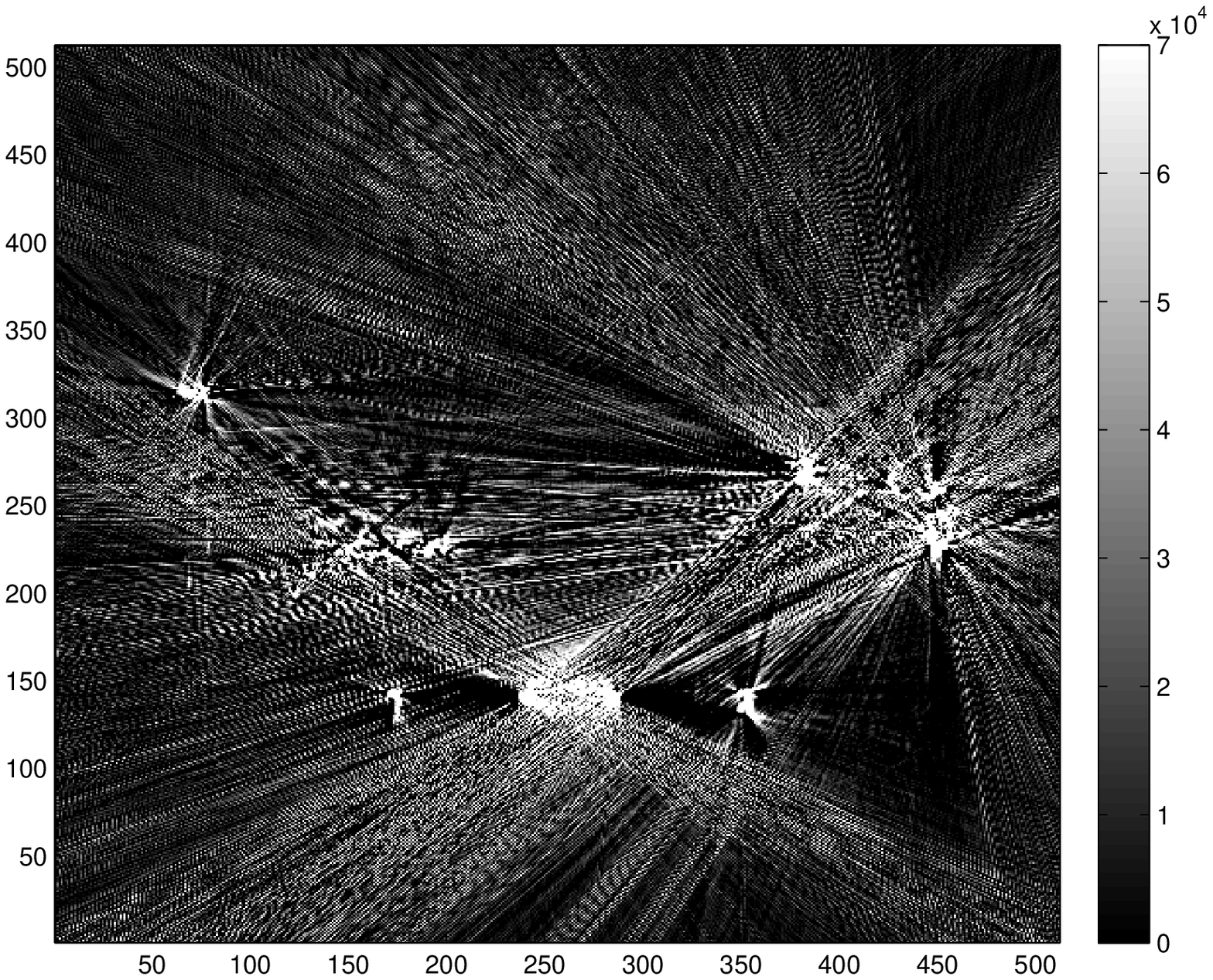}{(b)}
\includegraphics[scale=.3]{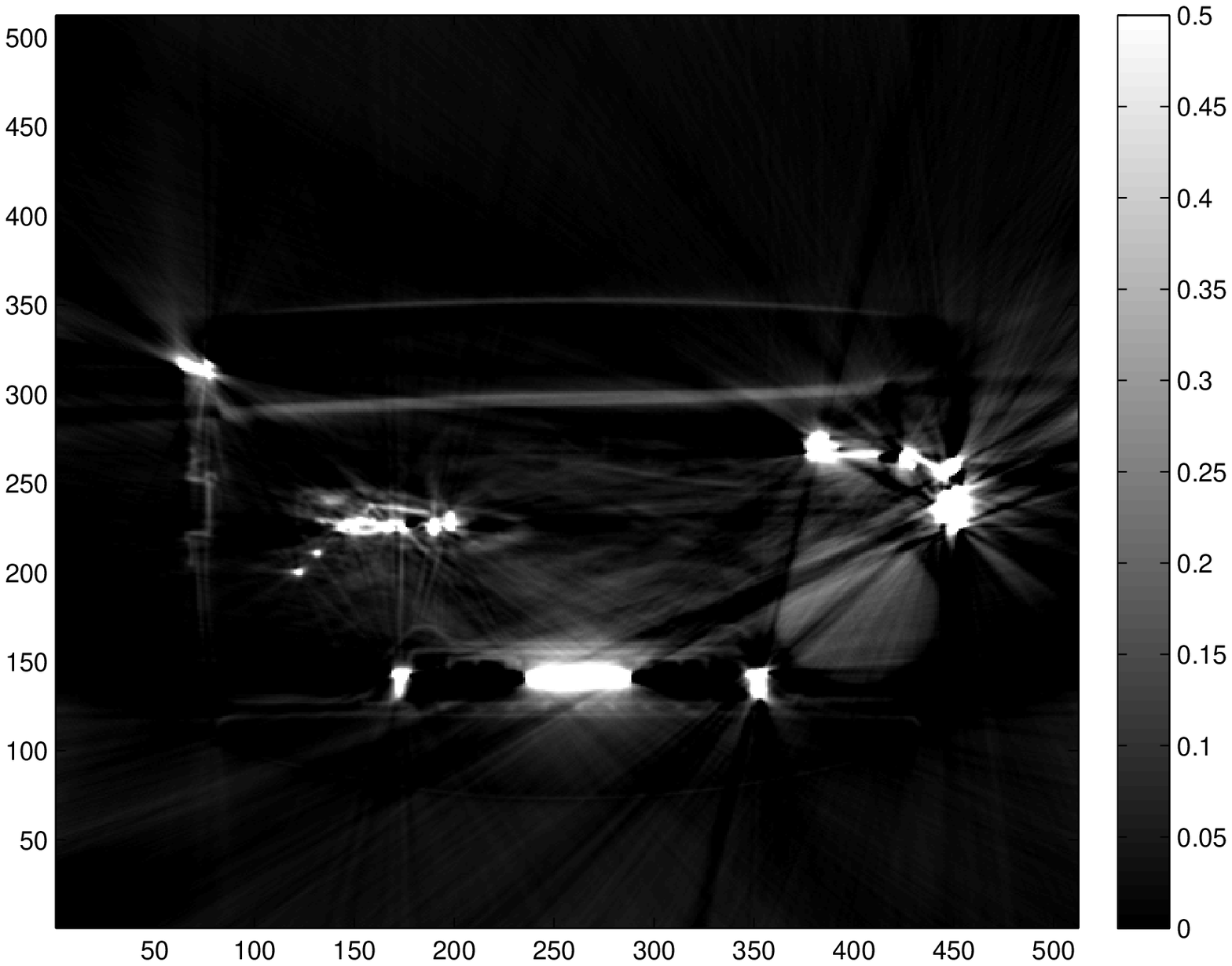}{(c)}
\includegraphics[scale=.3]{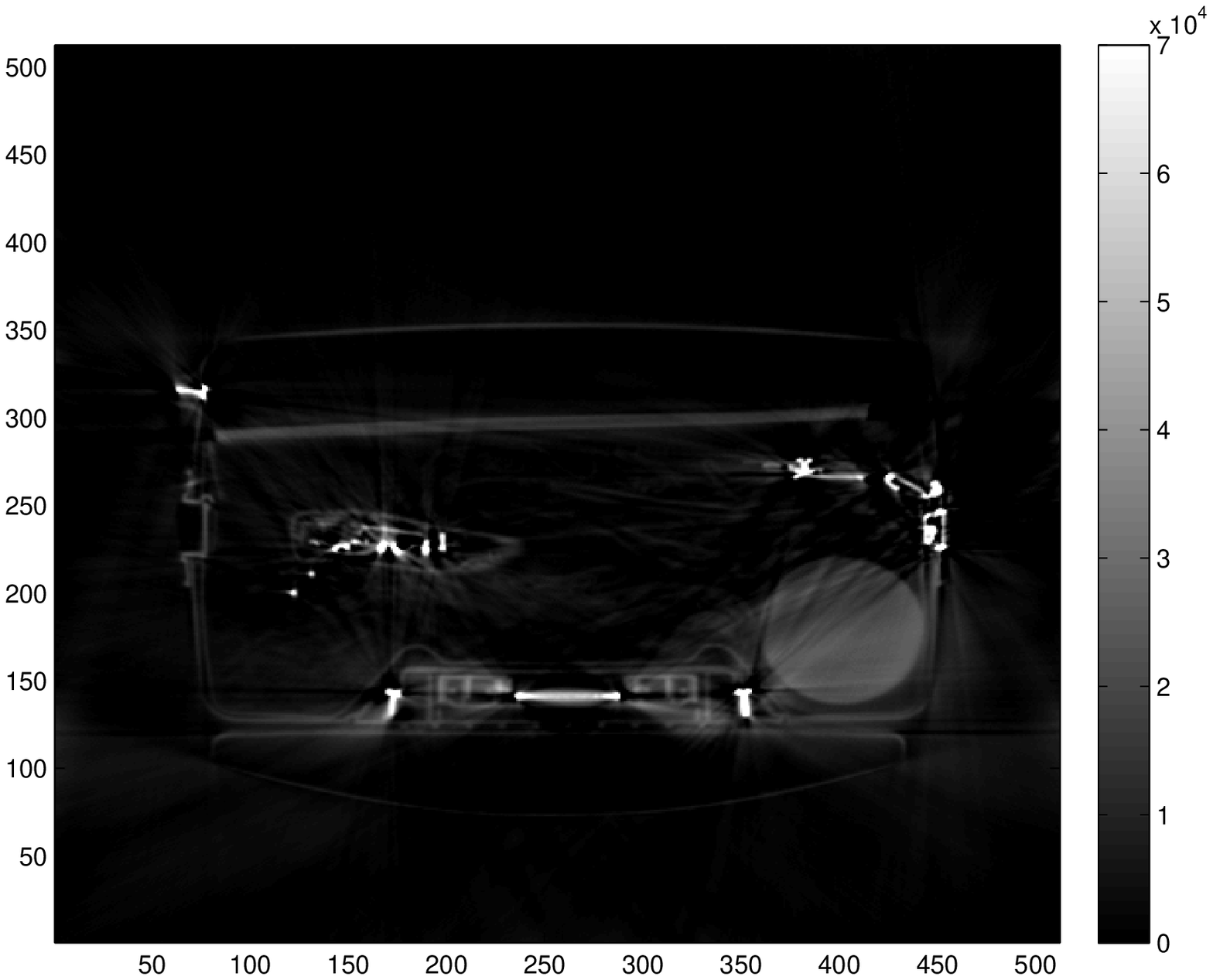}{(d)}
\includegraphics[scale=.3]{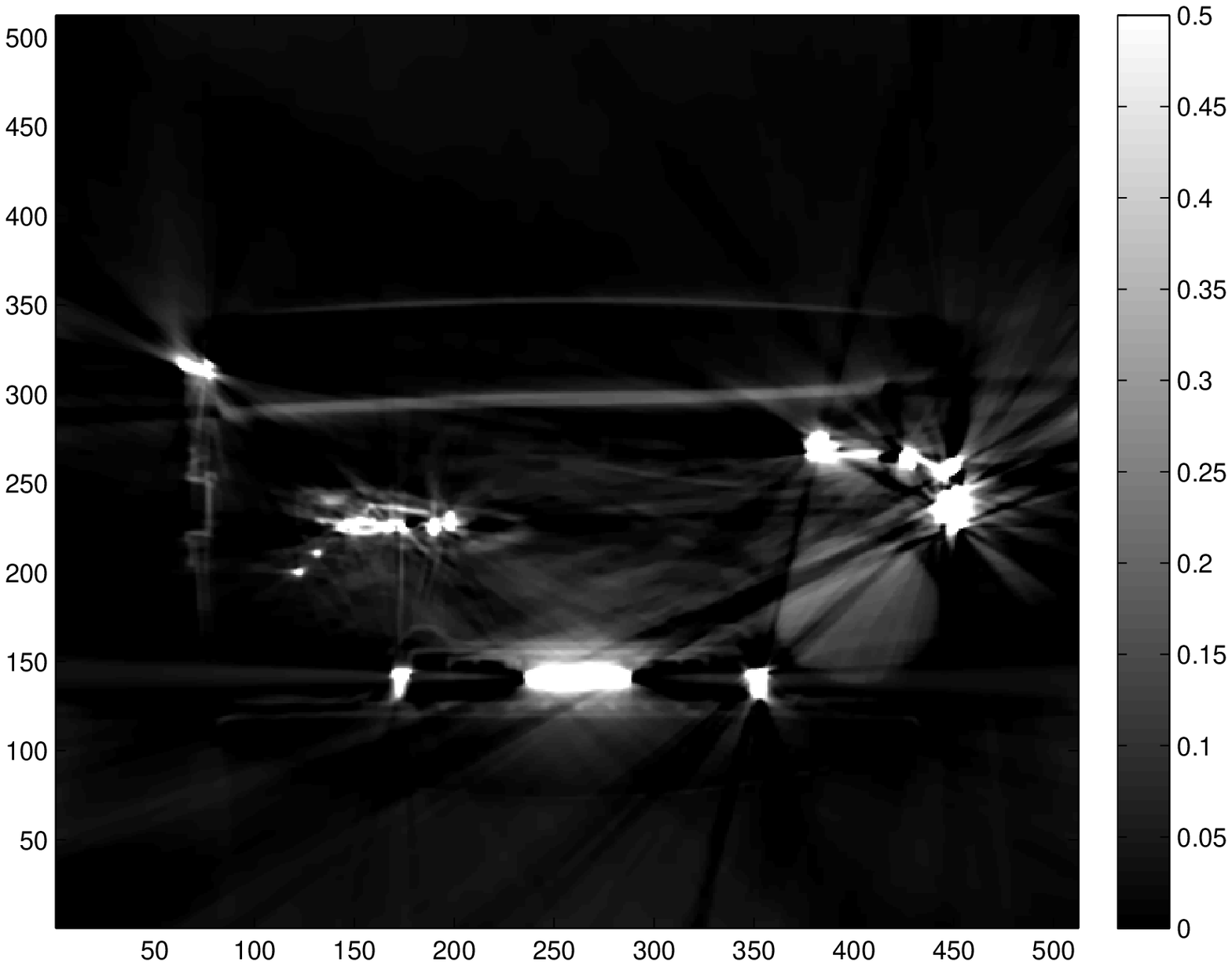}{(e)}
\includegraphics[scale=.3]{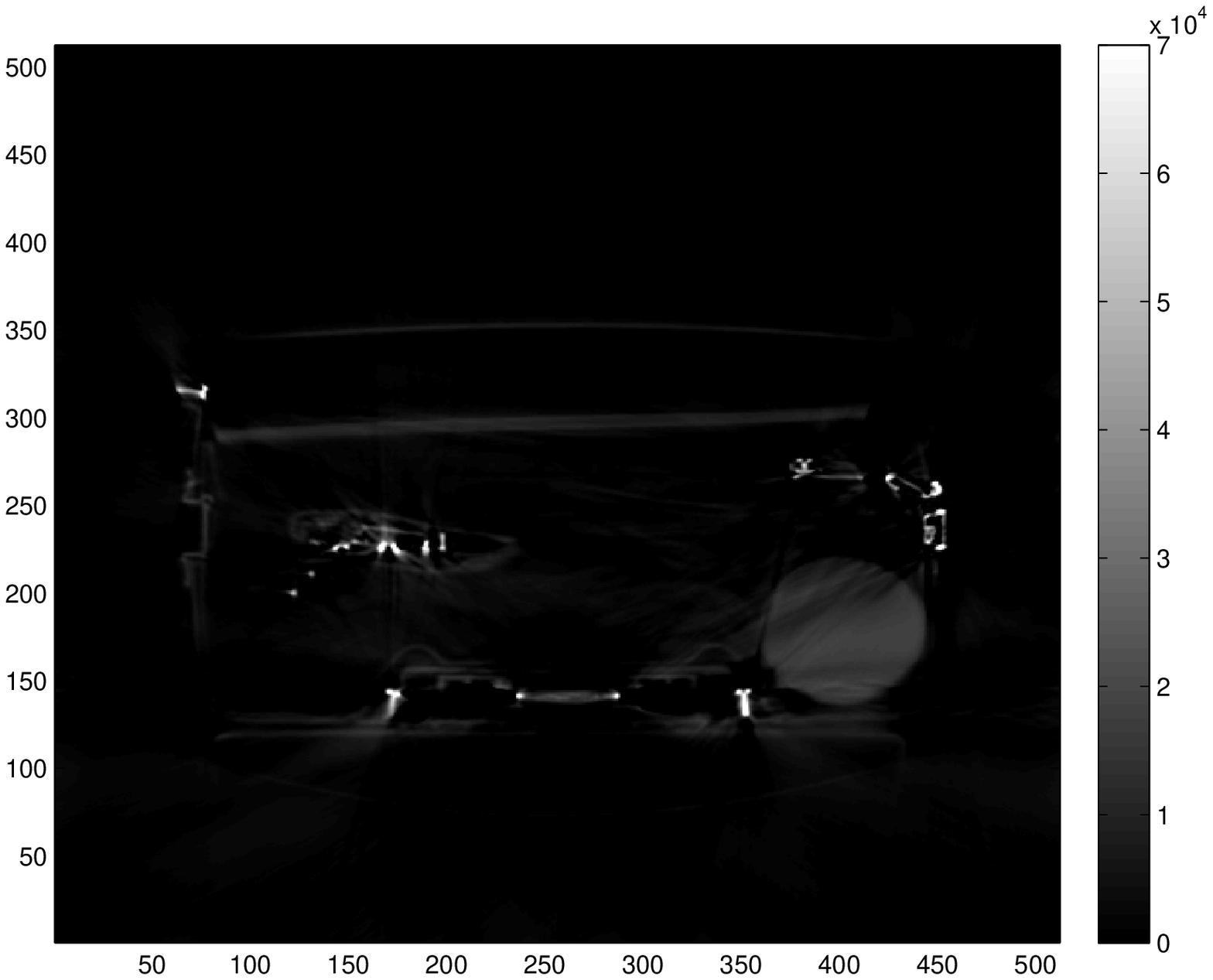}{(f)}
\caption{Packed suitcase including rubber sheet, water bottle, and shoe.  Compton images are on left, photoelectric on right.  Top row (a,b) are legacy; 2nd row (c,d) are iterative results but with no regularization; 3rd row (d,e) are iterative results with proposed regularization.} 
\label{fig:mc281}
\end{figure}

Figures~\ref{fig:MC195},~\ref{fig:mc38}, and ~\ref{fig:mc281} show three representative slices for different bags.     In each figure, Compton estimates are shown in the left column while photoelectric estimates on shown on the right.  The legacy YBC approach~\cite{ying} is shown on the top row; the middle row shows results from our method, but with weights for TV regularization of Compton and NLM regularization of photoelectric set to zero, to disable regularization.  Finally, the bottom row shows our proposed method with both regularization terms enabled.  For all results, the TV penalty assigned a low weight to avoid blurring of textured objects ($\lambda_{TV} = 0.01$) and the NLM parameters were tuned as discussed above.

While the legacy method of Ying \emph{et al.} does allow image reconstruction, the images (in particular photoelectric) are of relatively poor quality.  Moving to an iterative, polyenergetic solution (middle row) leads to noticeably improved estimates.  However, best estimates (reduced artifact, improved homogeneity of objects) are shown when the regularization terms are enabled.  These characteristic results are seen across a variety of scans, such as the densely packed suitcase in Fig.~\ref{fig:MC195}, the much less densely packed suitcase in Fig.~\ref{fig:mc38}, and the suitcase in Fig.~\ref{fig:mc281} which contains significant amounts of metal.

\begin{figure}[H]
\centering
\includegraphics[scale=0.3]{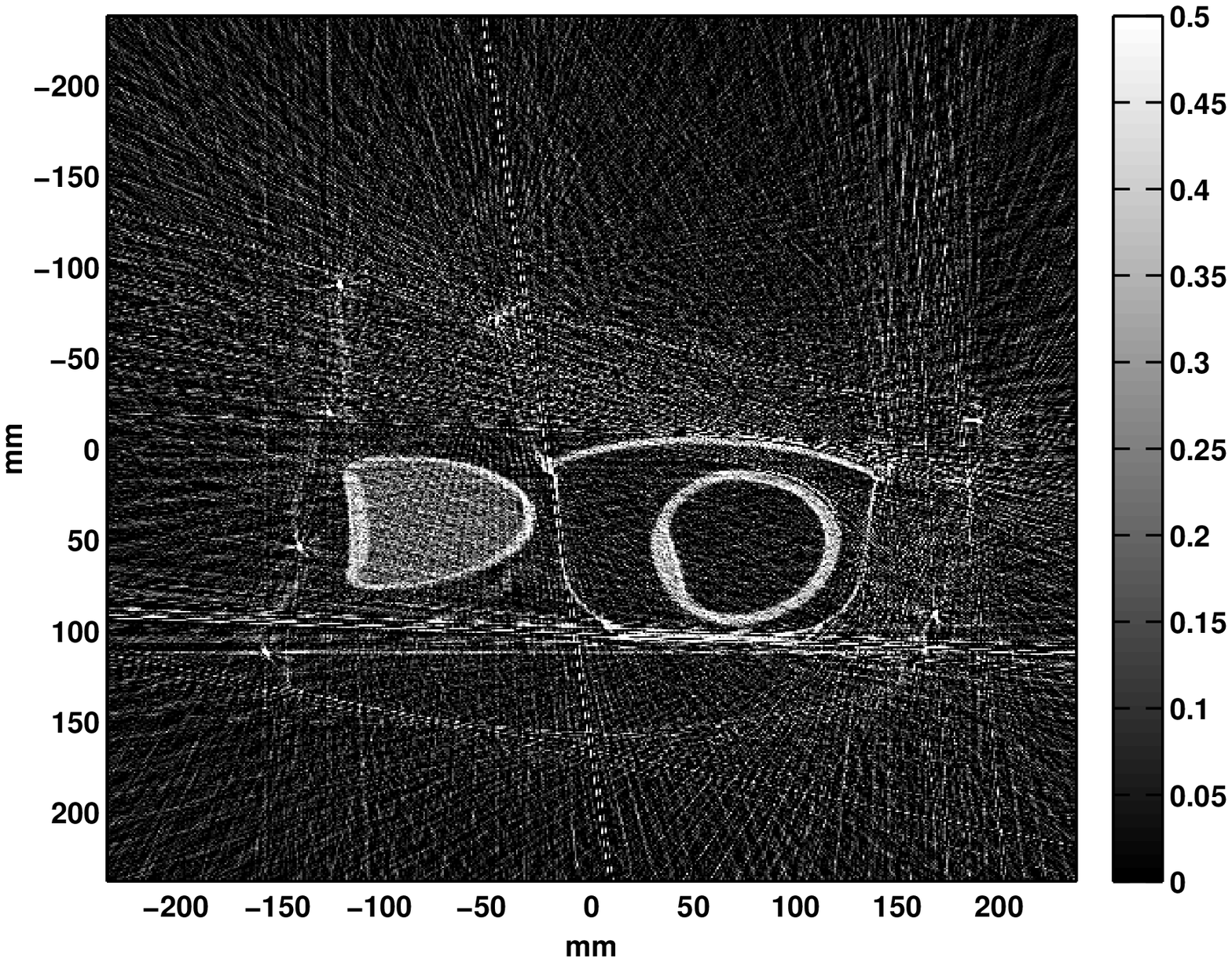}{(a)}
\includegraphics[scale=0.3]{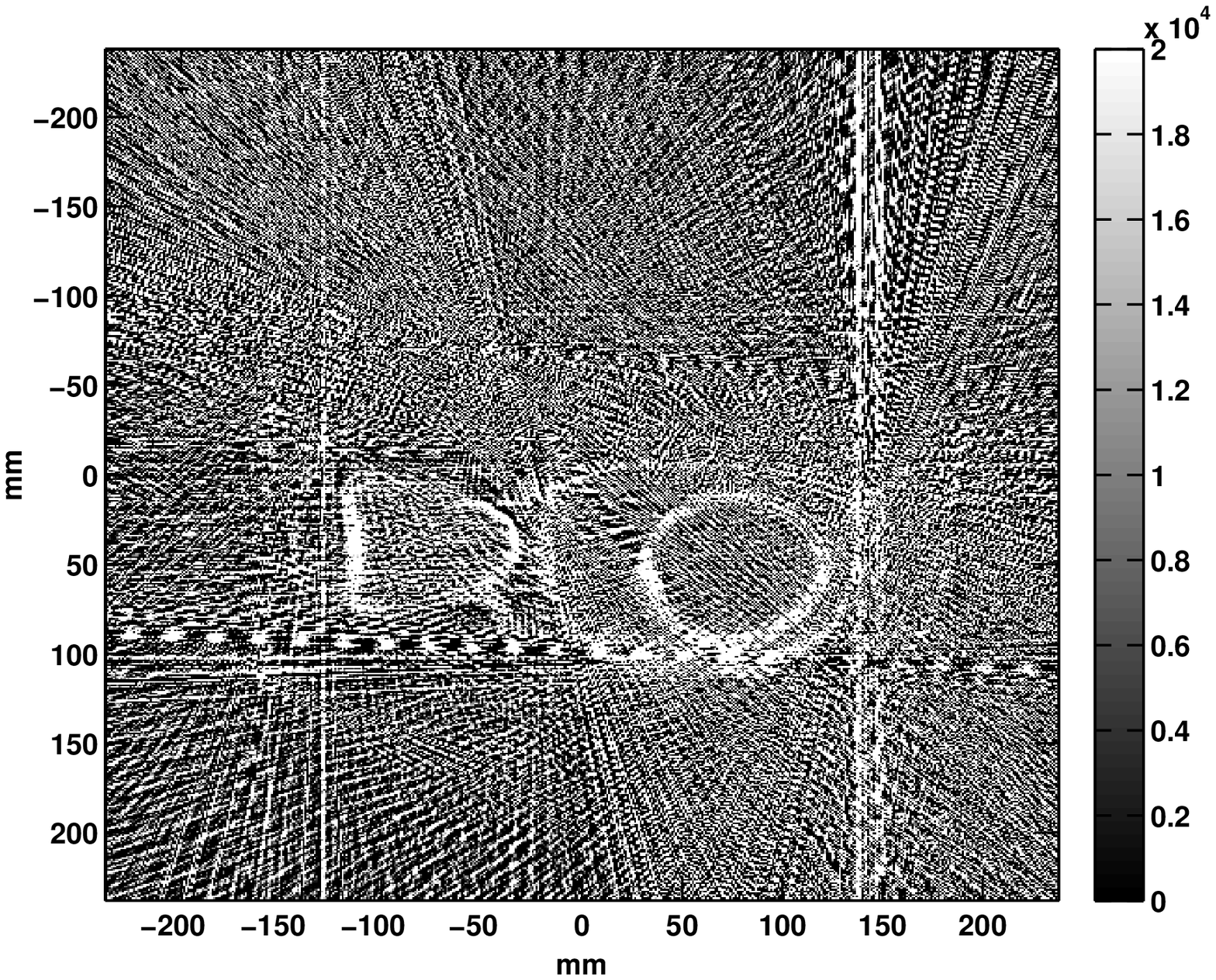}{(b)}
\includegraphics[scale=0.3]{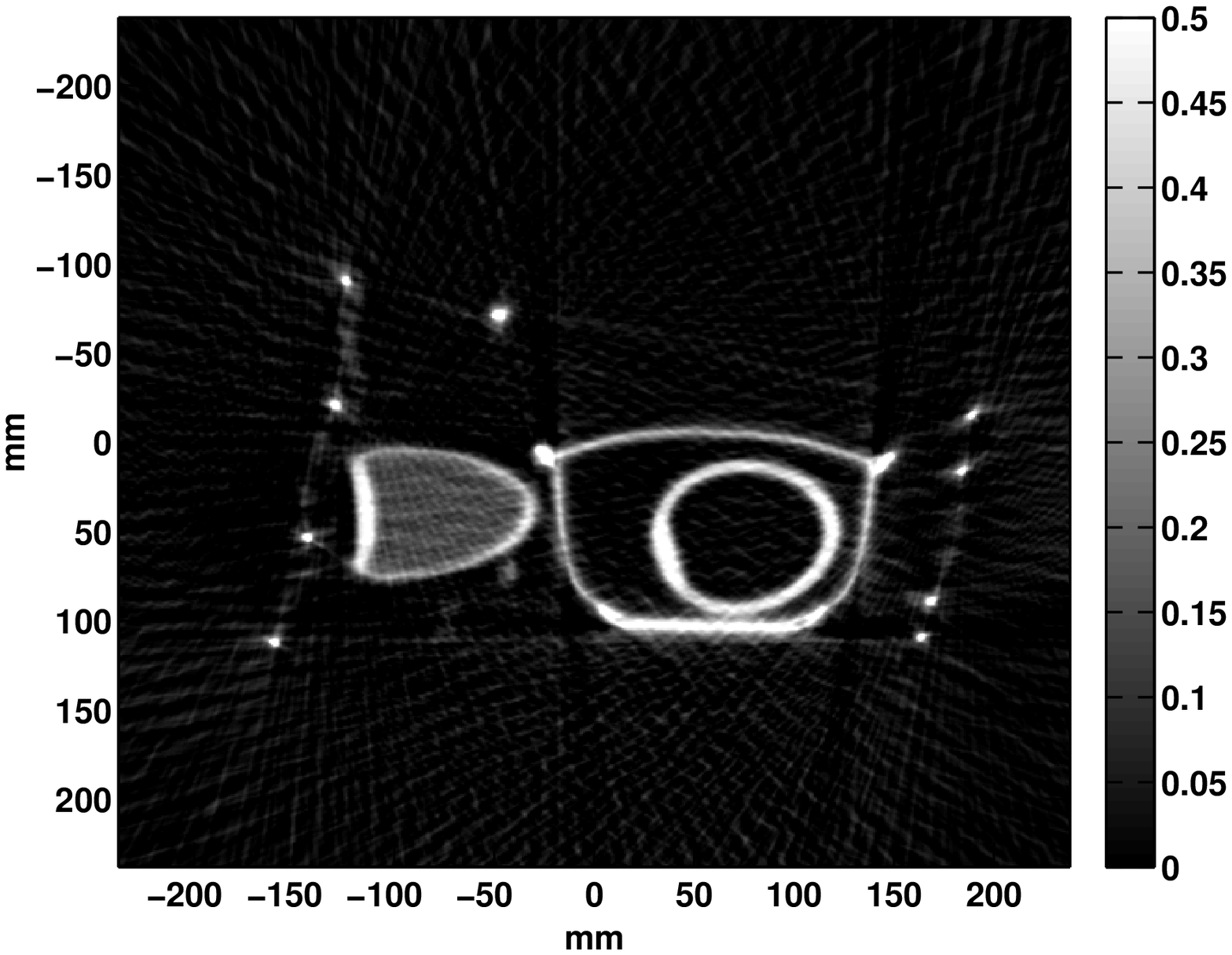}{(a)}
\includegraphics[scale=0.3]{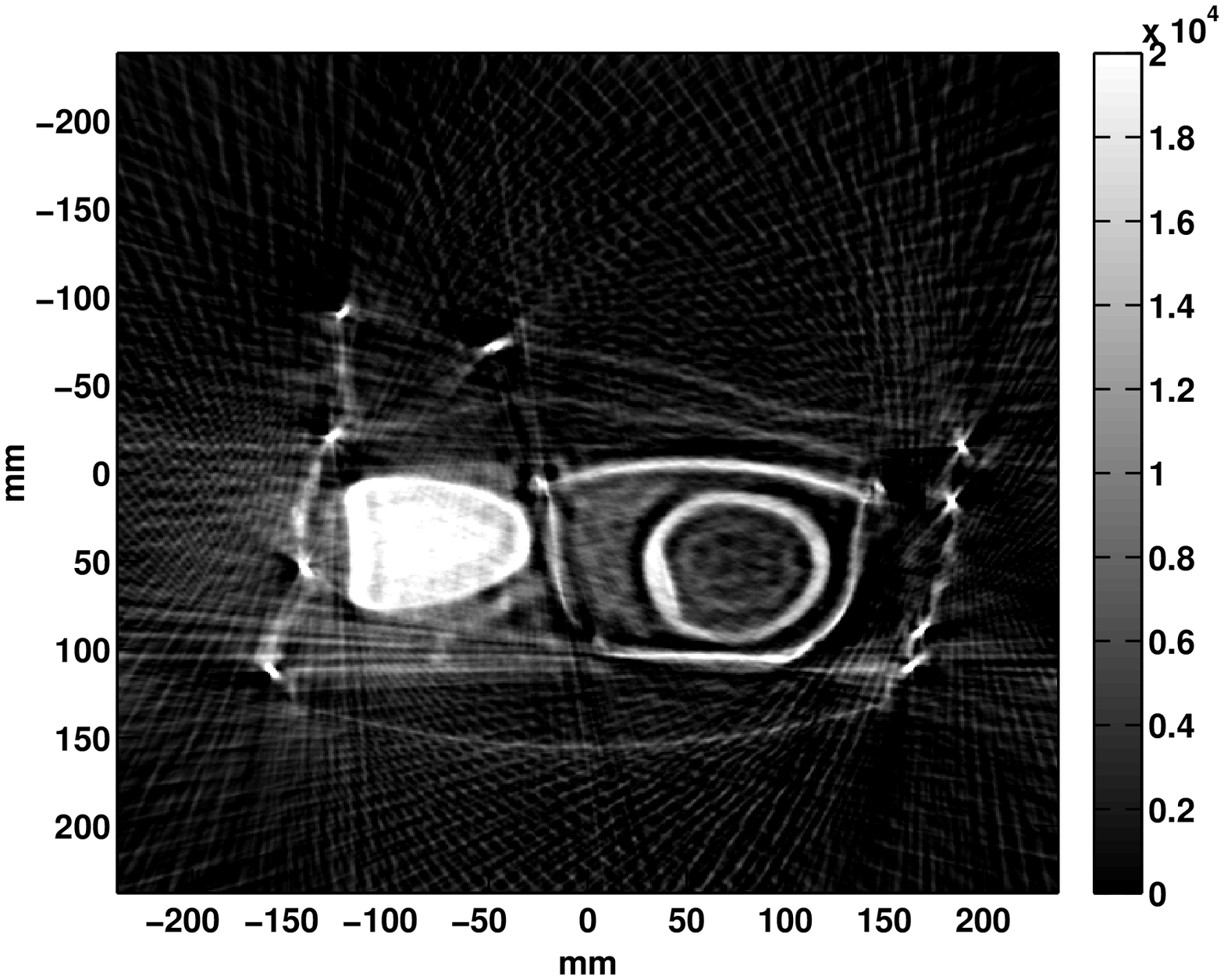}{(b)}
\includegraphics[scale=0.3]{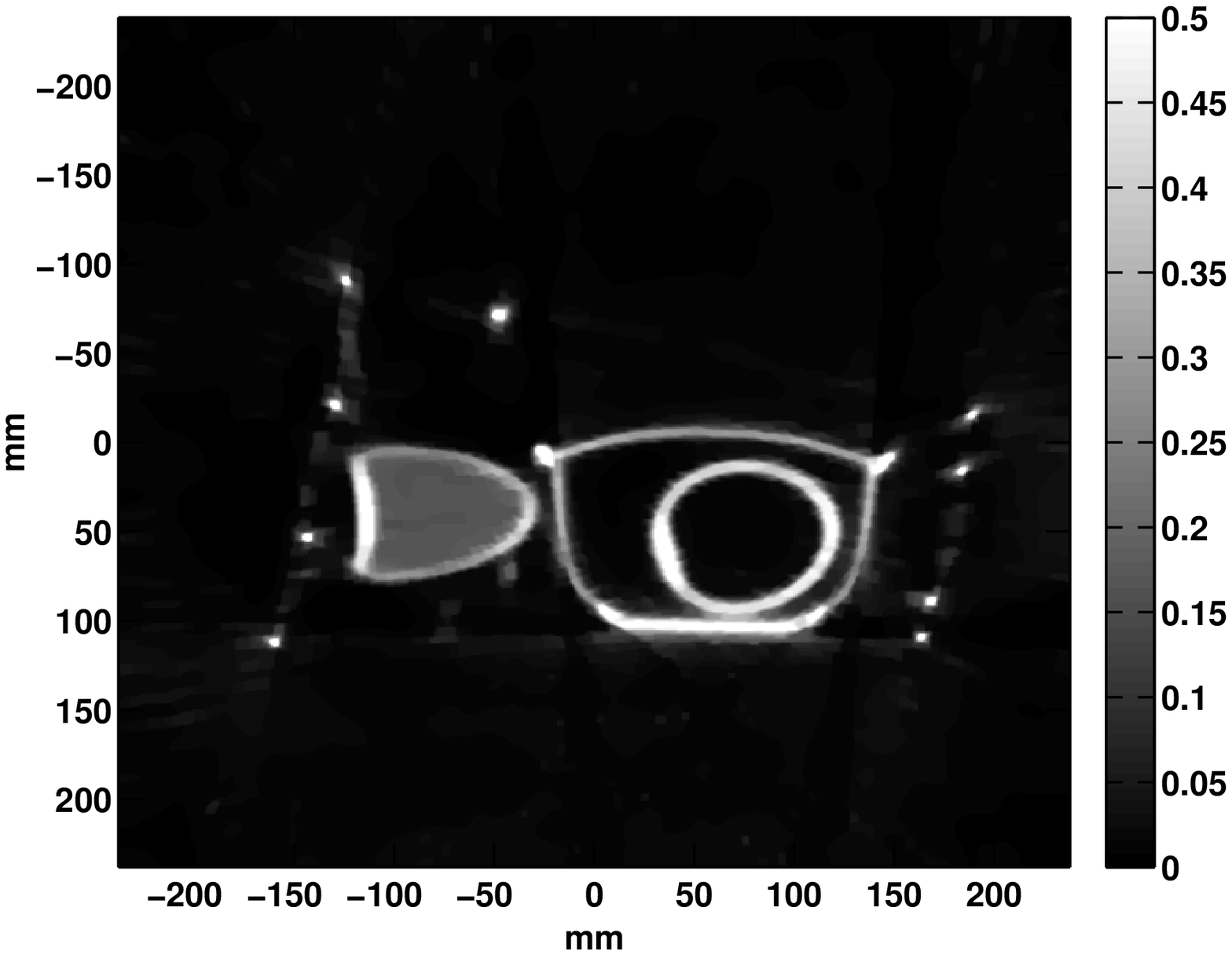}{(c)}
\includegraphics[scale=0.3]{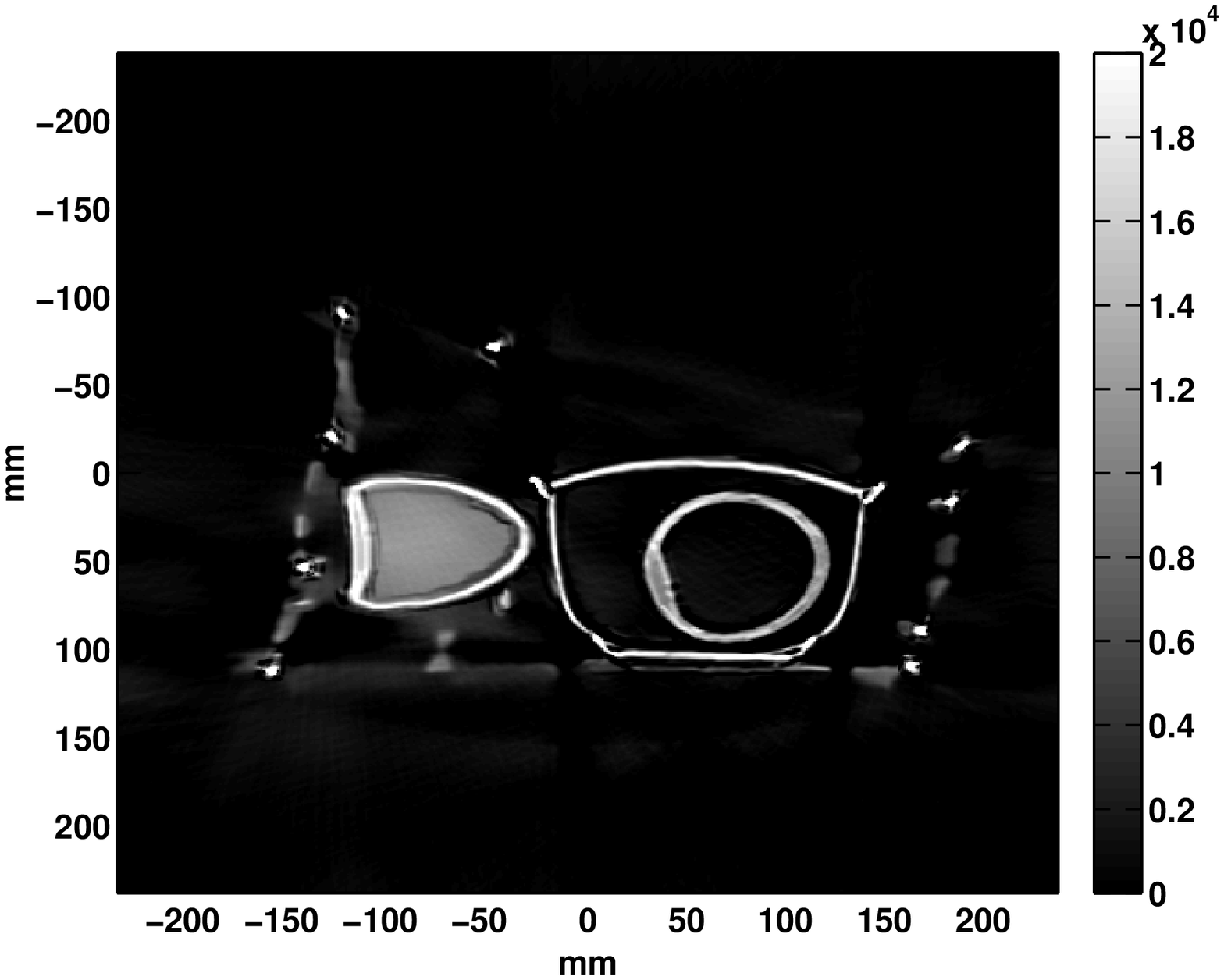}{(d)}
\caption{Packed suitcase including metal cooking pots and bottles, sinogram angles subsampled by factor of 10: Compton image, without TV (a) and with TV (b) ($\lambda_{TV}=0.01$).  Note  that TV removes some image noise.}
\label{fig:HCsubsampCompton}
\end{figure}

It is also interesting to consider the impact of regularization on  sparse-view  systems, in which scan times are reduced by scanning the object at fewer angles (or a limited angular range).   To test performance under this case, we subsampled the Imatron sinograms by selecting every $10^{th}$ angle, thus reducing the (rebinned, parallelized) sinogram from 720 angles to 72.  The system model was similarly decimated, and the reconstruction code was otherwise unchanged.  
Results are plotted in  Fig.~\ref{fig:HCsubsampCompton}.    Both the legacy and unregularized results are noticeably degraded by the reduced sampling.  The best image is recovered when both TV and NLM regularization are applied.  Improvements here are more noticeable than results shown above, in  part due to the improved efficacy of TV denoising on the Compton image.

\subsubsection{Uncertainty clouds for material estimates}

To provide quantitative measures of image improvement, manual segmentations of homogeneous test objects were created by researchers at Stratovan, Inc.   The three types of material selected were water, doped water, and rubber (neoprene sheets); examples of each type of material were pulled from separate scans. Examples of these objects include the rubber sheets shown in Figs.~\ref{fig:MC195} and ~\ref{fig:mc281} and the water containers seen in Figs.~\ref{fig:mc38} and ~\ref{fig:mc281}.

\begin{figure}[H]
\centering
\includegraphics[scale=0.3]{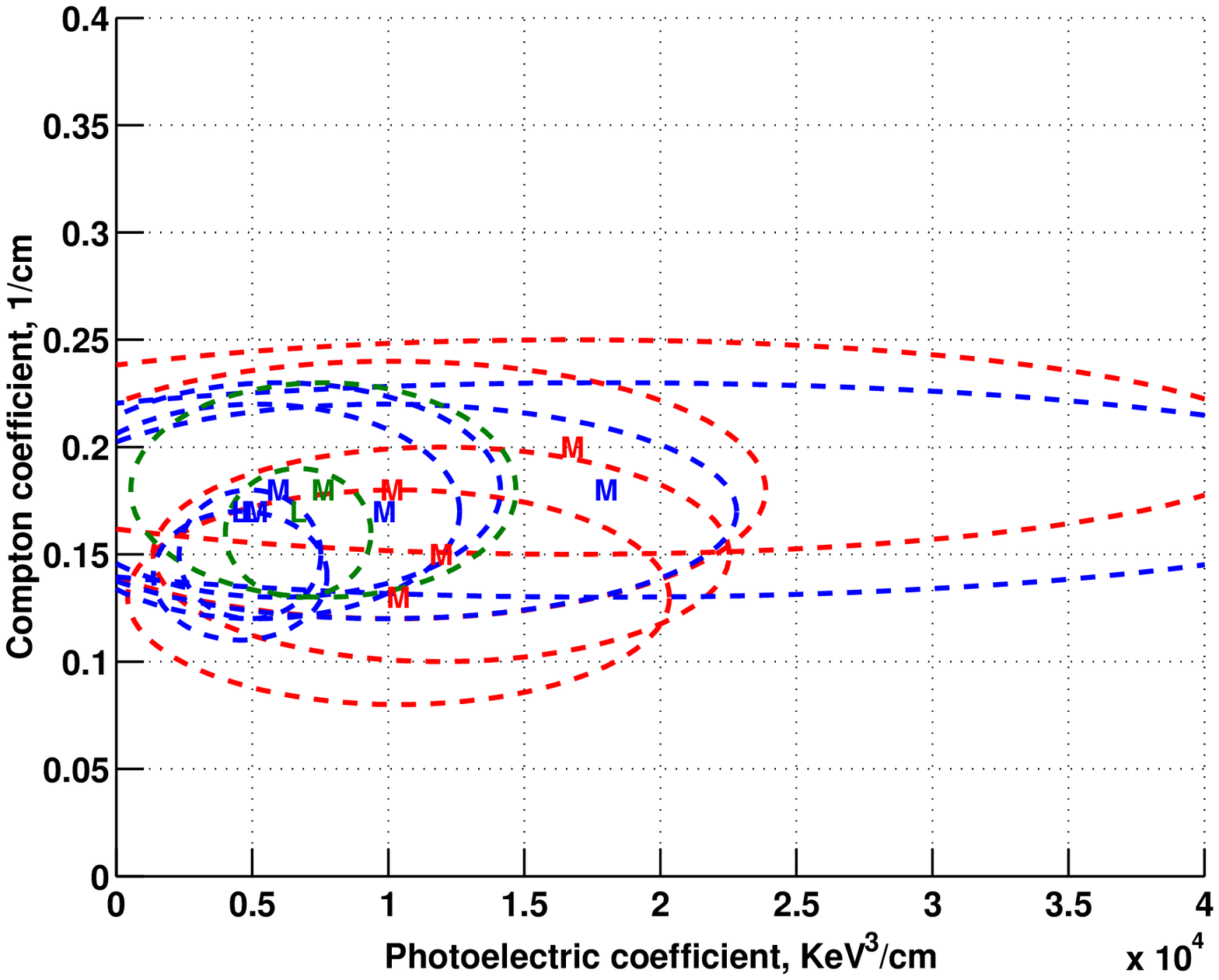}{(a)}
\includegraphics[scale=0.3]{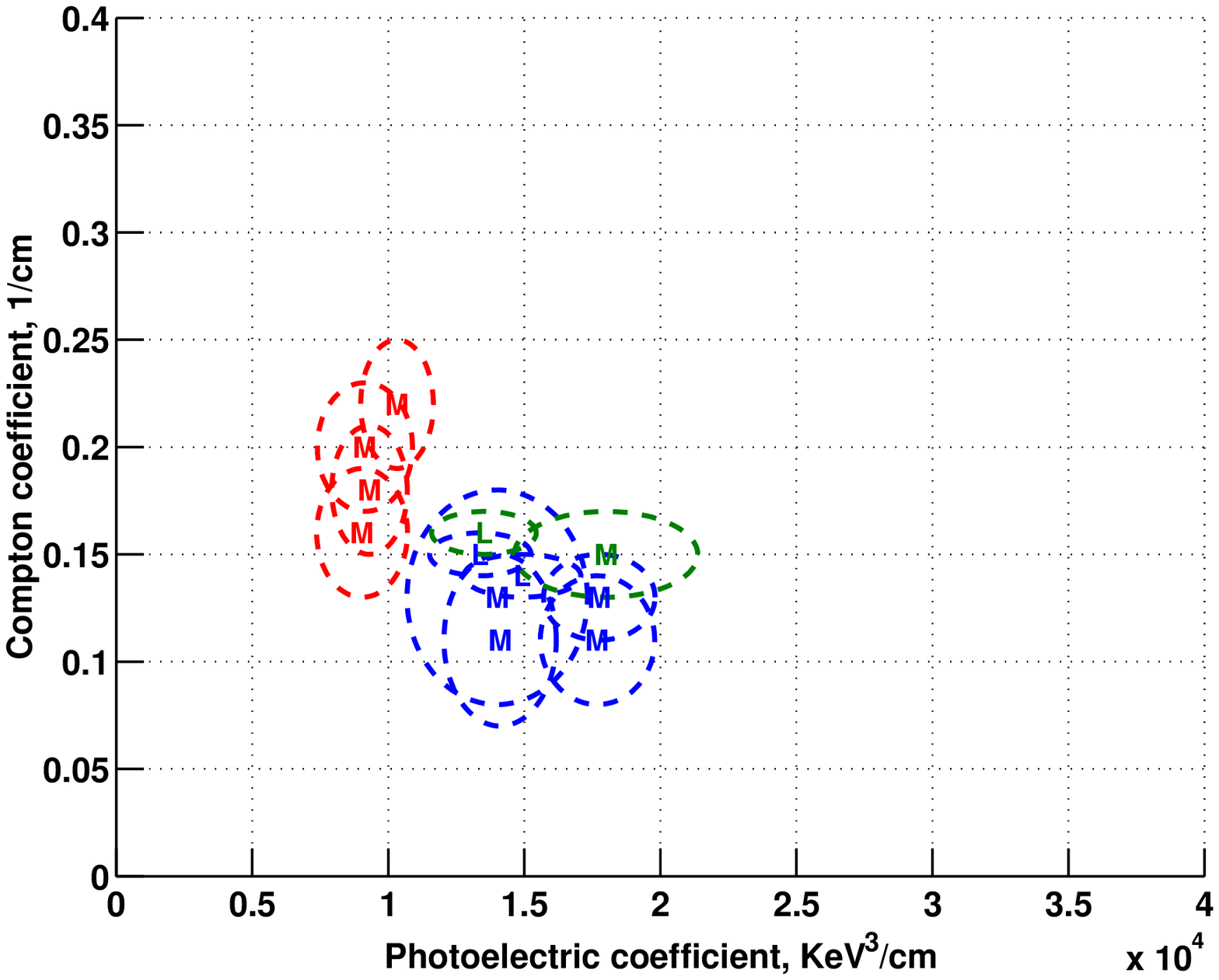}{(b)}
\caption{Material parameter estimates for manually-segmented objects.  For each object and each method, the average Compton/photoelectric value gives the ellipse centroid, along with ellipses extending out one standard deviation in each parameter.  Results are shown for (a) the legacy YNC method;  (b) 
the proposed ADMM solution including TV and NLM penalties.}
\label{fig:loClouds}
\end{figure}

For both the legacy YNC and our proposed method, the estimated Compton and photoelectric coefficients were extracted for all pixels within each of the manually segmented objects, and the mean and standard deviation were calculated.  Ideally, the mean values should be very repeatable for all objects made of the same material, and the standard deviation should approach zero as objects are homogenous.  The calculated mean and standard deviations were used to generate parameter ``clouds" for Compton and photoelectric estimates, as shown in Fig.~\ref{fig:loClouds}.  Results are shown for legacy results (a) and the approach developed here (b).  Here, each ellipse corresponds to one segmented object.  The object centroid is set by the mean Compton and photoelectric values in the object, while the ellipses extend to mean $\pm$ 1 standard deviation.  The material type is coded by color (water, doped water, and rubber).

In the legacy results (subfigure a), Compton is observed to have much lower variation about the mean than PE, as the standard deviations in estimated PE are very large.  
 Subfigure b) shows results obtained using our proposed method.   The PE image is stabilized, as reflected in lower PE standard deviations, and materials are much better clustered.   Visual inspection of the images shows that in some cases the Compton images are affected by streaking artifacts generated by metal; those particularly affect the water object in Fig.~\ref{fig:mc281}, which contributes the point that slightly overlaps the rubber region.   However,  material separation  is relatively good  and should only improve if metal artifact reduction~\cite{wang1996iterative,zhang2007reducing} was incorporated into processing.


\section{Discussion and Conclusions}
This paper has outlined an iterative image formation approach for dual-energy CT data.  The main contributions of this work include:
\begin{itemize}
\item In what we believe to be a novel contribution for dual energy CT, we have used patch-based regularization to stabilize the poorly estimated photoelectric image based on the much more stably estimated Compton image, and have demonstrated  improvements on both simulated and actual data.  We further demonstrated that using this approach in conjunction with Total Variation (TV) denoising of the Compton image leads to additional image improvements.
\item We exploited the convex nature of the patch-based regularization scheme, which allows us employ the recently developed Alternating Direction Method of Multipliers (ADMM) method, which can be used to create parallelizable implementations.
\item We compared our method to a state-of-the art sinogram decomposition method and show improved homogeneity and reduced noise, particularly in recovered photoelectric images.
\end{itemize}

We see several promising areas for future research.  First, 
we did not include metal artifact reduction (MAR) steps in our processing, and the effects of metal can be noticeable, for example in Fig.~\ref{fig:mc281}.  Our approach to regularizing the photoelectric image depends on good structural information being available from the Compton image, which is not a good assumption when metal artifacts are large.  Thus it would clearly be beneficial to combine methods previously developed for MAR~\cite{wang1996iterative,zhang2007reducing} with the reconstruction approaches outlined above, and we anticipate this would improve cases such as Fig.~\ref{fig:mc281}.
 Second, it may be possible to improve performance of pre-reconstuction decomposition methods such as Ying \emph{et al.}~\cite{ying} by including patch-based regularization in these techniques, which would help to address the noise problems seen with this approach.  Finally, it should be possible to apply these concepts to stabilize multi-energy CT images formed using  energy-discriminating detectors.

\section{Acknowledgments}   
 
This work was part of a multi-team effort funded by DHS through the ALERT Center at Northeastern University.   As such, we benefited greatly from the contributions of other team members.  The authors gratefully acknowledge helpful insights provided by Dr. Carl Crawford, Oguz Semerci, and Limor Martin,  Profs. Taly Gilat-Schmidt and Clem Karl, the groups of Profs. Charles Bouman and Ken Sauer for providing their calculated system matrix, and Stravovan, Inc. for manual segmentation results.  

This material is based upon work supported by the U.S. Department of Homeland Security, Science and Technology Directorate, under Task Order Number HSHQDC-12-J-00056.   The views and conclusions contained in this document are those of the authors and should not be interpreted as necessarily representing the official policies, either expressed or implied, of the U.S. Department of Homeland Security.


\bibliography{CTbib}   
\bibliographystyle{unsrt}  

\end{document}